\documentclass[sigconf]{acmart}
\usepackage{multirow}
\usepackage{tikz}
\usepackage{amsmath}
\usepackage{amsfonts}
\usepackage{color} 
\usepackage{mathtools}
\usepackage{amsthm}
\usepackage{xspace}
\usepackage{overpic} 
\usepackage{booktabs}
\usepackage{bbding}
\usepackage{colortbl}
\usepackage{pifont}
\usepackage{makecell}
\usepackage{hyperref}
\usepackage{pifont}
\usepackage{algorithm}  
\usepackage{algorithmic}  

\usepackage{balance}

\makeatletter
\DeclareRobustCommand\onedot{\futurelet\@let@token\@onedot}
\def\@onedot{\ifx\@let@token.\else.\null\fi\xspace}
\def\eg {\emph{e.g}\onedot} 
\def\ie{\emph{i.e}\onedot} 
 
\def\etc{\emph{etc}\onedot} 
 
\def\etal{\emph{et al}\onedot}
\makeatother

\AtBeginDocument{%
  }

\setcopyright{acmcopyright}
\copyrightyear{2018}
\acmYear{2018}
\acmDOI{XXXXXXX.XXXXXXX}

\acmConference[Conference acronym 'XX]{Make sure to enter the correct
  conference title from your rights confirmation emai}{June 03--05,
  2018}{Woodstock, NY}
\acmPrice{15.00}
\acmISBN{978-1-4503-XXXX-X/18/06}

\begin{document}

\title{A Unified Membership Inference Method for Visual Self-supervised Encoder via Part-aware Capability}


\author{Jie Zhu$^{1,2}$, Jirong Zha$^3$, Ding Li$^{1,2}$, Leye Wang$^{1,2}$}
\authornote{Corresponding author}
	\renewcommand{\authors}{Jie Zhu, Jirong Zha, Ding Li, Leye Wang}
	\renewcommand{\shortauthors}{Jie Zhu, Jirong Zha, Ding Li, Leye Wang}
\affiliation{%
    $^{1}$\institution{Key Lab of High Confidence Software Technologies (Peking University), Ministry of Education \country{China}} 
	$^{2}$\institution{School of Computer Science, Peking University, Beijing, China}
  $^3$\institution{Shenzhen International Graduate School, Tsinghua University, Shenzhen, China}
   \country{}
  }
\email{zhujie@stu.pku.edu.cn, zhajirong23@mails.tsinghua.edu.cn, {ding_li, leyewang}@pku.edu.cn}

\begin{abstract}
Self-supervised learning shows promise in harnessing extensive unlabeled data, but it also confronts significant privacy concerns, especially in vision. In this paper, we aim to perform membership inference on visual self-supervised models in a more realistic setting: \textit{self-supervised training method and details are unknown for an adversary when attacking as he usually faces a black-box system in practice.} In this setting, considering that self-supervised model could be trained by completely different self-supervised paradigms, \eg, masked image modeling and contrastive learning, with complex training details, we propose a unified membership inference method called PartCrop. It is motivated by the shared part-aware capability among models and stronger part response on the training data. Specifically, PartCrop crops parts of objects in an image to query responses with the image in representation space. We conduct extensive attacks on self-supervised models with different training protocols and structures using three widely used image datasets. The results verify the effectiveness and generalization of PartCrop. Moreover, to defend against PartCrop, we evaluate two common approaches, \ie, early stop and differential privacy, and propose a tailored method called shrinking crop scale range. The defense experiments indicate that all of them are effective. Our code is available at \url{https://github.com/JiePKU/PartCrop}.

\end{abstract}


\keywords{Visual self-supervised learning, membership inference, part-aware capability}

\received{20 February 2007}
\received[revised]{12 March 2009}
\received[accepted]{5 June 2009}

\maketitle

%

\section{Introduction}

Self-supervised learning~\cite{chen2020simple, chen2020big} is proposed recently and quickly attracts much attention due to its excellent capability of learning knowledge from substantial unlabeled data. 
While promising, self-supervised models are likely to involve individual privacy during training as they see massive data~\cite{chen2020simple} (\eg, personal or medical images) that are possibly collected from website without personal authorization~\cite{twitter_stop_url, FTC_settlement}. Twitter has required company Clearview AI to
stop using public images from its platform for model training~\cite{twitter_stop_url}. 
In addition, visual self-supervised models are usually used for services where users can feed their data and use its exposed API to extract data features (Internet giants such as Google and Amazon are already
offering ``machine learning as a service”~\cite{liu2021encodermi}.). It increases the privacy risk as adversaries~\footnote{In the rest of the paper, without incurring the ambiguity, we will use `adversary' to indicate the person who manipulates attack and use `attacker' to indicate the attack model an adversary utilizes.} may act as a normal person and feed some elaborate data that could lead to output containing sensitive privacy information. Then adversaries can analyze these outputs and make privacy-related inference. 

In this research, we are interested in performing membership inference (MI) against self-supervised image encoders. Specifically, given an image, our goal is to infer whether it is used for training a self-supervised image encoder~\footnote{We primarily focus on membership inference in a black-box setting~\cite{Shaow_Learning, ResAdv} where only the output from the encoder is available as we deem that this setting is more likely to meet the realistic situation.}. Prior to our research, Liu~\etal have customized a membership inference method called EncoderMI~\cite{liu2021encodermi} for contrastive learning~\cite{caron2021emerging, he2020momentum}. However, self-supervised learning encompasses methods beyond contrastive learning, \eg, masked image modeling which has recently gained great prevalence~\cite{he2022masked, chen2022context}. Our experiment in Tab~\ref{tab:baseline} shows that EncoderMI fails to attack models trained by masked image modeling. More importantly, EncoderMI assumes adversaries know the training details and employs the same data augmentations and hyperparameters as the target encoder. \textit{We contend that this setting is unjustified, as it might not accurately represent the true situation in real-world scenarios.}

In practice, self-supervised models are typically treated as black boxes~\cite{liu2021ml, Shaow_Learning, pre_conf_ML_Leaks, ResAdv}, particularly for internet giants like Google. In other words, we are only aware that the model within the black box is trained using one of the self-supervised methods, without knowledge of the specific method and its details~\cite{Shaow_Learning}~\footnote{In~\cite{Shaow_Learning}, it is said ``The details of the models and the training algorithms are
hidden from the data owners.''. We give a more detailed explanation in Sec~\ref{sec:background} Part 3.}. Hence, the practical application of EncoderMI may be limited, and we summarize two underlying reasons:

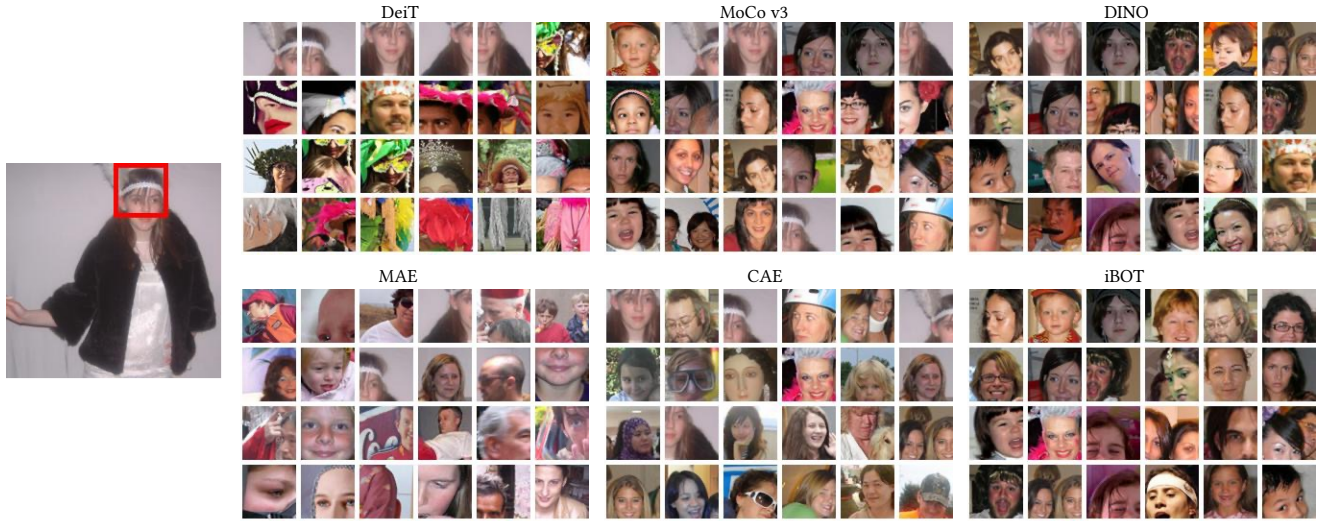
\begin{figure*}[t]
	\footnotesize
	\centering
  \vskip -0.05in
        \begin{overpic}[width=.98\linewidth]{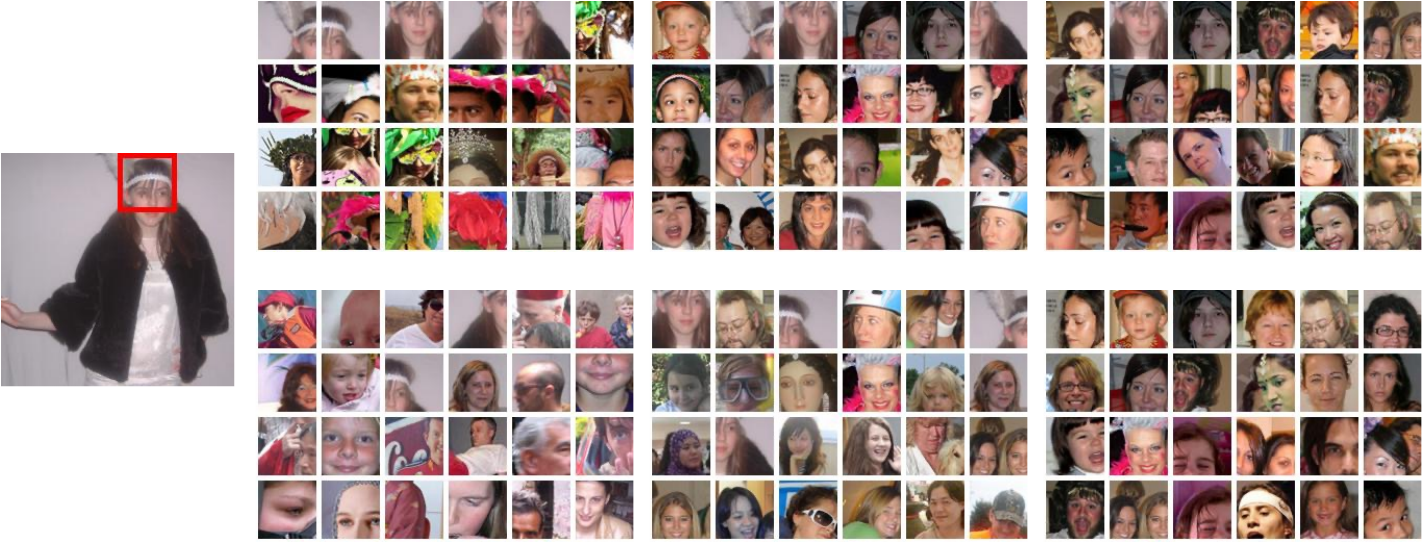}
    \put(28.7,38.4){DeiT}
    \put(54.5,38.4){MoCo v3}
    \put(83.8,38.4){DINO}
    \put(28.5,18.2){MAE}
    \put(56.6,18.2){CAE}
    \put(83.8,18.2){iBOT}
	\end{overpic}
 \vskip -0.1in
	\caption{DeiT~\cite{touvron2021training} uses supervised learning. MAE~\cite{he2022masked} and CAE~\cite{chen2022context} are masked image modeling based methods.  DINO~\cite{caron2021emerging} and MoCo v3~\cite{chen2021empirical} are contrastive learning based methods. iBOT~\cite{zhou2021ibot} combines the two paradigms. This figure is borrowed from ~\cite{zhu2023understanding}. We refer readers of interest to \cite{zhu2023understanding}.} 
	\label{fig:part}
 \vskip -0.15in
\end{figure*}

$\bullet$ \textit{Firstly,  EncoderMI relies on the prior of contrastive learning.} During training, contrastive learning mandates image encoder to minimize the distance between representations of two randomly augmented views of the same image, establishing a crucial prior where feature vectors of these views are similar. However, this prior is not always established, as different self-supervised paradigms often bestow encoders with distinct characteristics. For instance, masked image modeling is a generative method that reconstructs masked parts of an image using the visible parts, encouraging the encoder to prioritize local context within the image. This difference leads to EncoderMI's failure in Tab~\ref{tab:baseline} for masked image modeling.

$\bullet$ \textit{Secondly, the training recipe of self-supervised model is usually unknown in reality.} In the  field of self-supervised learning, current methods can be categorized into different families, \eg, contrastive learning~\cite{caron2021emerging, he2020momentum}, masked image modeling~\cite{he2022masked, chen2022context}, clustering~\cite{caron2020unsupervised}, information maximization~\cite{zbontar2021barlow, bardes2021vicreg, ermolov2021whitening}, \etc. In real-world scenarios, adversaries often face a black-box system~\cite{Shaow_Learning, pre_conf_ML_Leaks} (contain a self-supervised model) and actually do not know the specific method used and training details~\cite{Shaow_Learning}.

An intuitive solution to this circumstances is to systematically evaluate membership inference methods designed for different visual self-supervised methods and adopt the best-performing one. This would help adversaries overcome the limitation of lacking knowledge about the self-supervised training recipe. Though intuitively effective, we have found, to the best of our knowledge, there are no existing membership inference methods designed for visual self-supervised methods other than EncoderMI. This lack of alternatives can be attributed to the longstanding dominance of contrastive learning in the field of self-supervised learning, while a more promising paradigm, \ie, masked image modeling~\cite{he2022masked, chen2022context}, has only recently emerged.  
Moreover, even if we assume the existence of several inference methods, the process of enumerating them may be considerably intractable and inefficient due to the various pipelines and data processing involved, wasting substantial time.









\textbf{To mitigate this dilemma, we aim to propose a unified membership inference method without prior knowledge of how self-supervised models are trained.} 
This setting involves two folds: First, self-supervised models could be trained by different paradigms, \eg, contrastive learning or masked image modeling~\footnote{In the field of self-supervised learning, contrastive learning~\cite{chen2020simple, he2020momentum, chen2021empirical, caron2021emerging} is the most representative method and masked image modeling~\cite{bao2021beit, he2022masked} is the most prevalent method recently and widely used by companies, \eg, MicroSoft~\cite{xie2022simmim}, Facebook~\cite{he2022masked}, \etc. Hence, they constitute our primary focus, while we also assess other three representative (or newly proposed) paradigms, illustrating the generalization of our method, see Sec~\ref{unified}};  Secondly, training details, \eg, data augmentations and hyperparameters, are also unknown for adversaries. In this difficult circumstance, to attack successfully, we propose to find a key characteristic meeting the requirements: \textit{\ding{172} The characteristic is common among these self-supervised models. This helps us disregard the training discrepancy in methods and details. \ding{173} The characteristic is expected to be more salient for training data than test data due to many epochs of training. This helps us distinguish training and test data.}  

\begin{figure}
	\centering{\includegraphics[width=1\linewidth]{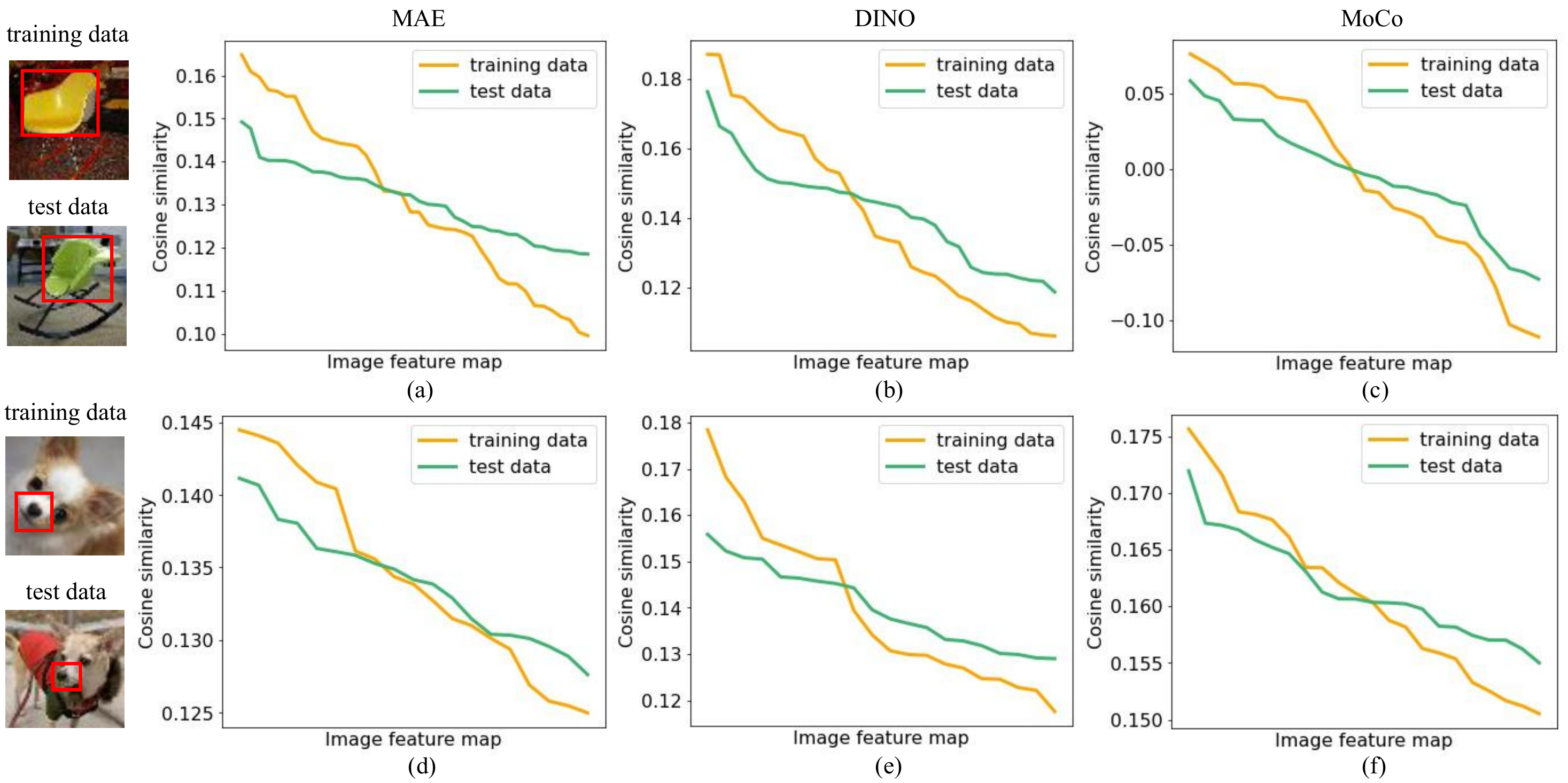}}
	\caption{Part response visualization on MAE (masked), DINO (contrastive), and MoCo (contrastive). Images are from Tinyimagenet~\cite{tinyimagenet_le2015tiny}. (a), (b), (c) Similarity curves of the chair image and chair seat part on MAE, DINO, and MoCo, respectively. (d), (e), (f) Similarity curves of the dog image and dog muzzle part on MAE, DINO, and MoCo, respectively.}
	\label{fig:vis}
    \vskip -0.1in
\end{figure}

Interestingly, we find a characteristic (we refer to as ``part-aware capability") exactly meeting the requirements. This capability enables self-supervised models to capture characteristics of different \textbf{parts of an object~\footnote{The part is not simply a patch of an image. It means a part of an object, \eg, the head of a person, the mouth of a dog, and the eye of a cat.}}, producing discriminative part representations. As shown in Fig~\ref{fig:part}, in a part retrieval experiment, the cropped part in red boundingbox (\ie, a human head) is used to query in a large image dataset. We find that contrastive learning (DINO and MoCo v3) and masked image modeling (MAE and CAE) basically retrieve various human heads while supervised method (DeiT) retrieves other parts, \eg, hair, ear, mouth, and clothes. This indicates that these self-supervised methods are superior at part perception (meet \ding{172}). This is also verified by concurrent work~\cite{zhu2023understanding}. Simultaneously, we are the first to find that training data exhibits a stronger part-aware capability compared to test data by showing their part response. 
In Fig~\ref{fig:vis}, for each image containing a chair or dog, we manually crop it to obtain the \textit{part} (chair seat or dog muzzle) in the red box and resize it to suitable size. Next, we calculate the cosine similarity between each vector of the feature map of entire image and the part feature vector from the encoder, sorting the scores in descending order. Notably, the similarity curves for the training data are generally steeper than test data, indicating that part representations of training data are more discriminative than that of test data (meet \ding{173}).




Motivated by this capability, we propose to use both global image and cropped parts where global image serving as a measurement platform gauges cropped parts' response. 
A strong response means the image is likely from training set. However, manually cropping the part of an object in an image is precise but inefficient, while automatic cropping requires sophisticated design. Hence, we adopt a roundabout strategy by randomly cropping numerous image patches to potentially yield as part-containing crops as possible. These patches, to some extent, can be considered as parts, and along with the global image, they are fed into a black-box self-supervised model to produce corresponding features. 
Afterward, part features serve as queries and interact with the image feature map to generate part-aware response. This response is regarded as \textit{membership feature} following~\cite{liu2021encodermi} and is collected as input of a simple attacker for membership inference. We term this membership inference method as \textbf{\textit{PartCrop}}. Note that PartCrop is a \textit{unified} method as we do not specify the training methods and details of the given self-supervised model. 
Empirically, although the ``part" may not be as precise as manually cropped parts, PartCrop still significantly improves performance compared to random guessing (This may be the real situation when adversaries are unaware of the training methods and details.). Finally, to defend against PartCrop, we evaluate two widely-used methods, \ie, early stop and differential privacy, and propose a tailored defense method called shrinking crop scale range.

Our contribution can be summarized as follows:

$\bullet$ To the best of our knowledge, our work is the first to perform membership inference on self-supervised models without training recipe in hand. It is a more valuable setting in real life.

$\bullet$ To the best of our knowledge, we are the first to leverage the part-aware capability in self-supervised models for membership inference by proposing PartCrop. In brief, PartCrop crops parts as queries to measure the responses of image feature map and collect them as membership feature to perform inference.
 
$\bullet$ To defend against PartCrop, we evaluate two common approaches, \ie, early stop and differential privacy, and propose a tailored method called shrinking crop scale range.   

$\bullet$ Extensive experiments are conducted using three self-supervised models with different training protocols and structures on three computer vision datasets. The results verify the effectiveness of the attack and defense method.

\section{Background}\label{sec:background}

\textbf{Membership Inference (MI).} Prior research~\cite{carlini2019secret, nasr2019comprehensive} point out that a neural model is inclined to memorize the training data. Such characteristic makes the model exhibit different behavior on training data (\ie, members) versus test data (\ie, non-members)~\cite{hu2022membership}. For example, a well-trained neural model could give a higher confidence (or stronger response) to training data than test data. Based on this observation, membership inference~\cite{Shaow_Learning, ResAdv} is proposed to infer whether a data record is used for training or not. When it comes to sensitive data, personal privacy is exposed to great risk. For example, if membership inference learns that a target user's electronic health record (EHR) data is used to train a model related to a specific disease (\eg, to predict the length of stay in ICU~\cite{ma2021distilling}), then the adversary knows that the target user has the disease. In this work, we aim to perform membership inference on self-supervised models, go further toward a more reasonable setting where an adversary does not know self-supervised training recipe, and propose a unified attack method dubbed as PartCrop. 

\textbf{Self-supervised Learning.} Self-supervised learning (SSL)~\cite{chen2020simple} is proposed to help learn knowledge from substantial unlabelled data. Current approaches in self-supervised learning can be categorized into different families, among which masked image modeling~\cite{bao2021beit, he2022masked, chen2022context, zhou2021ibot} and contrastive learning~\cite{chen2020simple, he2020momentum, grill2020bootstrap, chen2021empirical, caron2021emerging} are the most prevalent and representative. Hence, we mainly consider them in this work and give a brief introduction below: \textbf{Masked image modeling (MIM)~\cite{he2022masked} is a generative method.} Usually, it contains an encoder and a decoder. The encoder takes the masked image as input and outputs the features. Afterwards, the encoded features are fed into the decoder aiming to reconstruct the masked parts (\eg, RGB color) in raw images. Once finishing pretraining, only the encoder is preserved for downstream tasks, \eg, image classification. The part-level reconstruction process enables the model to gradually become sensitive to image parts; \textbf{Contrastive learning (CL)~\cite{he2020momentum} is a discriminant method.} It asks an encoder to pull the representations of two random views cropped from the same image close while pushing away the representations from different images.  After pretraining, the encoder is used to extract features for downstream tasks. The crops from an image usually contain part of an object. Maximizing the discrepancy of crops from different images makes part representations discriminative.

\textbf{Unknowing the self-supervised training recipe for an adversary is a reasonable setting.} In reality, service systems are typically well-packaged, like black boxes~\cite{pre_conf_ML_Leaks, Shaow_Learning, ResAdv, liu2021ml}. Users input data and receive output after payment, without knowledge of the underlying model or its structure~\cite{Shaow_Learning}. This service mode is commonly used and widely adopted by internet giants such as Google and Amazon~\cite{liu2021ml}. This manner allows these companies to offer their services through simple APIs, thus making machine
learning technologies available to any customer~\cite{Shaow_Learning, ResAdv}. More importantly, it helps protect their core technologies from being stolen by adversaries~\cite{liu2021ml}. In the case of a self-supervised model system, the training recipe - which includes the self-supervised method used, data augmentation strategy, and hyperparameters - is critical for producing a successful self-supervised model for service. Therefore, it cannot be exposed to adversaries. Hence, when attacking self-supervised models, adversaries typically face a black box and are unaware of the used self-supervised method and details. From this view, this is a reasonable representation of the real-world setting.

\section{Threat Model}

For adversaries' capability, we consider a more realistic setting where adversaries have no knowledge of the target SSL model, \eg, training details, and only have black-box access to it. For the threat model, following~\cite{liu2021ml}, we mainly consider two kinds of threat models for evaluation, \ie, \textbf{Partial} model and \textbf{Shadow} model. They are trained under the corresponding settings.


\textbf{Partial} setting means that an adversary may obtain partial training and test data of a target dataset, \eg, 50\%, due to data leakage, such as in the Facebook scandal~\cite{facebook_url}. In this situation, we assume that the adversary can directly use the partial data to train an attacker, and then attack the remaining data for evaluation.
  

\textbf{Shadow} setting indicates that an adversary has no knowledge of the training and test data of a target dataset. In this situation, the adversary has to resort to public datasets. Specifically, the adversary can first train a self-supervised model on a public dataset and then use this model and the public dataset to train an attacker. Afterward, the adversary uses this attacker to attack the target dataset. For example, the adversary can train an attacker on CIFAR100 and then perform an attack on a self-supervised model trained on Tinyimagenet. 

\textbf{Remark.} We regard attackers trained under partial setting as partial models (shadow models are the same). Considering that there could be numerous partial models (or shadow models) for different datasets, for simplicity, we use partial setting and shadow setting to indicate the type of threat model in experiments below. And we use partial setting if not specified.

\begin{figure*}[h]	\centering{\includegraphics[width=1\linewidth]{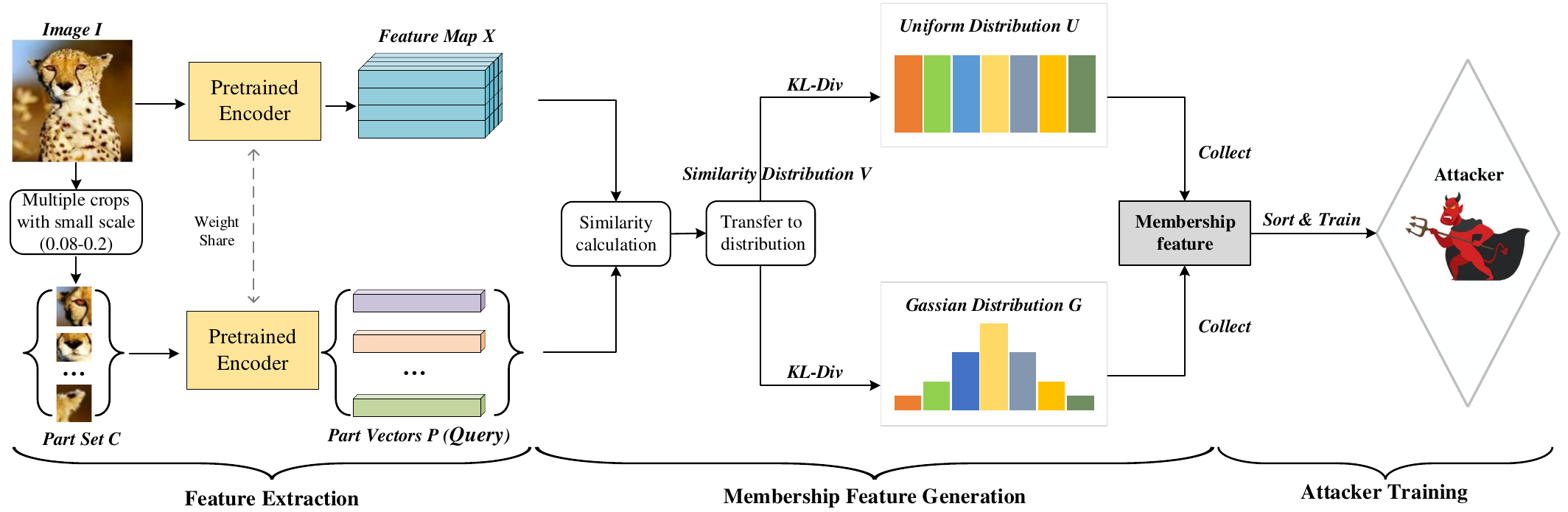}}
	\caption{An overview of PartCrop.}
	\label{overview}
 \vskip 0.05in
\end{figure*}

\section{Method}
\textit{\textbf{Overview.}} PartCrop contains three stages including feature extraction, membership feature generation, and attacker training as illustrated in Fig~\ref{overview}. Stage 1 is to extract image and part feature via mapping them to feature space through a shared self-supervised encoder. Then, the extracted image and part representations are used to generate membership features in Stage 2. Finally, we use the generated membership features to train a fully connected neural attacker in Stage 3. Note that in this work, we primarily focus on image data and leave the application of PartCrop to other domains as future work.

\textit{\textbf{Stage 1: Feature Extraction.}} Feature extraction is the first stage that contains two branches mapping image and part to the same representation space via a shared self-supervised encoder.

In the first branch, we extract the image feature. The whole image $I$ is fed into self-supervised model $\mathcal{F}$ that outputs a feature map $\chi \in \mathbb{R}^{H \times W \times D} $. The process is formulated as:
\begin{equation}
	\chi = \mathcal{F}(I) \,, \quad \chi \in \mathbb{R}^{H \times W \times D}
\end{equation}
$D$ is feature dimension. $H$ and $W$ are spatial size. We flatten the feature map along spatial dimension and reshape it as $\chi \in \mathbb{R}^{ N\times D}$ where $ N = H \times W $. 

In the second branch, we extract the part feature. We firstly randomly crop $m$ image patches that are likely to contain object parts from the whole image with crop scale $s$. We collect these crops in a set and denote it as $C = \{c_{0}, c_{1}, c_{2}, ......, c_{m-1}\}$ where $c_{i}$ is the $i$-th part crop. These crops then are fed into model $\mathcal{F}$ followed by avgpool and mapped to feature vectors in the same feature space as $\chi$. We denote these feature vectors as $P = \{p_{0}, p_{1}, p_{2}, ......, p_{m-1}\}$. We simply formulate it as:
\begin{equation}
P = avgpool(\mathcal{F}(C))
\end{equation}
For each element $p_{i}$ in $P$, $p_{i}$ is a feature vector with size $\mathbb{R}^{D}$. 

In our experiments, we set $m$ to $128$ and crop scale $s$ is set to $(0.08, 0.2)$ by default. We find that this setting is effective.

After finishing extracting features $\chi \in \mathbb{R}^{ N\times D}$ and $P = \{p_{0}, p_{1}, \\ p_{2}, ......, p_{m-1}\}$ are sent to the second stage.

\textit{\textbf{Stage 2: Membership Feature Generation.}}
Usually, in part-level perception, training data react more strongly than test data due to overfitting. Thus, in this stage, we regard each part crop representation vector $p_{i}$ in $P$ as a query to the feature map $\chi$ and return the response energy. And we consider the response energy as membership feature.

Given a query $p_{i}$, to obtain the response energy, we firstly calculate the similarity between query vector $p_{i}$ and feature map $\chi$ using the following formula:
\begin{equation} \label{eq:similarity}
\begin{split}
v_{i} = \chi \times p_{i} \quad \quad \quad \quad  \\ p_{i} \in \mathbb{R}^{D}\,, \; \chi \in \mathbb{R}^{ N\times D}\,, \; and \; v_{i} \in \mathbb{R}^{N}
\end{split}
\end{equation}
$\times$ indicates matrix multiplication. 

Now $v_{i}$ is the similarity vector between query $p_{i}$ and feature map $\chi$. An intuitive idea is to return the maximum value in $v_{i}$ as response energy. But this manner only focuses on local peak and ignores global similarity information. It could make membership feature not robust. Another method is to return the whole similarity vector. However, considering that we have $m$ queries, this could be somewhat over-size ($m\times N$) with potential redundance and wastes computational memory and resource for training an attacker.

Hence, we propose to leverage KL-divergence to integrate global similarity information.
Specifically, we transfer the global similarity vector to a similarity probability distribution following~\cite{vaswani2017attention} by using a softmax function:
\begin{equation}
\begin{split}
& v_{i}  = Softmax(v_{i}) \,, \\
& v_{ij} = \frac{e^{v_{ij}}}{\sum_{k=0}^{k=N-1} e^{v_{ik}}} \,, 
\end{split}
\end{equation}
where $v_{ij}$ is the $j$-th similarity value in $v_{i}$.

Then, to extract the similarity distribution characteristic, KL-divergence naturally stands out as it is designed to measure the difference between two distributions. Consequently, we use two common and independent distributions, \ie, uniform distribution $u_{i}$ and gaussian distribution $g_{i}$~\footnote{The gaussian distribution is randomly generated for each query. So we add a subscript $i$ to $u$ and $g$ for consistency. $u_{i}\sim U(0, D)$ and $g_{i}\sim N(0,1)$.}, as benchmarks, to evaluate the similarity distribution. The two distributions are easily implemented and morphologically different, introducing greater diversity in KL-divergence calculation. Empirically, we believe that the more the number of benchmark is, the richer the depicted similarity distribution characteristic is. In our experiment, we find that the two distributions are sufficient to achieve satisfying results.

In the end, our response energy $e_{i}$ for query $p_{i}$ is given by:
\begin{equation} \label{eq:response_energy}
\begin{split}
& e^{u}_{i} = KL\text{-}Div(v_{i} \; u_{i}) = \sum_{j=0}^{N-1} u_{ij}\log \frac{u_{ij}}{v_{ij}}   \,, \\
& e^{g}_{i} = KL\text{-}Div(v_{i}, \; g_{i}) = \sum_{j=0}^{N-1} g_{ij}\log \frac{g_{ij}}{v_{ij}} \,, \\
& e_{i} = [e^{u}_{i},\; e^{g}_{i} ] \,, 
\end{split}
\end{equation}
where $v_{ij}$, $u_{ij}$, and $g_{ij}$ are $j$-th  probability value in $v_{i}$, $u_{i}$, and $g_{i}$ respectively. 

As shown in Eq.~\ref{eq:response_energy}, our response energy $e_{i}$ contains two parts. The first is called uniform energy $e^{u}_{i}$ and the second is called gaussian energy $e^{g}_{i}$.

In our concrete implementation, we calculate the response energy for all queries in parallel. In this way, our method becomes more efficient and we can obtain the whole response energy $E = [E^{u},\; E^{g}],\; E \in \mathbb{R}^{ m\times 2}$ at once. Finally, we regard the acquired response energy $E$ as membership feature and use it to train a fully connected neural attacker in the next stage. 

\textit{\textbf{Stage 3: Attacker Training.}}
We use the extracted membership feature $E = [E^{u},\; E^{g}],\; E \in \mathbb{R}^{ m\times 2}$ to train an attacker $f$. It is an extremely simple fully connected network containing two MLP layers with ReLU as the activation function. 
We can also build a more complex attacker, \eg, add more layers and use self-attention module~\cite{vaswani2017attention}. However, our experiments indicate that a simple attacker performs well.  

To train the attacker, we firstly sort the uniform energy $E^{u} \in \mathbb{R}^{m}$ and gaussian energy $E^{g} \in \mathbb{R}^{m}$ by descending order respectively. Then, we concatenate $E^{u}$ and $E^{g}$ together and reshape it to a vector $E \in \mathbb{R}^{2m} $ as input. Finally, the attacker's output and corresponding label are fed into a loss function $\mathcal{L}$ for optimization. 

\begin{algorithm}[t]  
	\footnotesize
	\caption{PartCrop Procedure}  
	\label{alg:Framwork}  
	\begin{algorithmic}[1]  
		\REQUIRE A self-supervised model $\mathcal{F}$; An attcker $f$; A loss function $\mathcal{L}$; A training set $Tr$ (member) and a test set $Te$ (no-member); the number of part crops $m$ and the crop scale $s$;
		\ENSURE  
		A well-trained attcker $f$;
		\STATE \textcolor{gray}{\# load a minibatch $x$ with B images, half in $Te$, half in $Tr$.}  
		\STATE \textbf{for} $x$, $y$ in loader($\{Te,Tr\}$,$\{0,1\}$) \textbf{do}: 
		\STATE \quad $\chi$ = $\mathcal{F}(x)$ \quad \textcolor{gray}{\# The shape of $\chi$ : B $\times$ D $\times$H $\times$ W}
		\STATE \quad $\chi$ = flatten($\chi$, dim=($2,\;3$)) \quad \textcolor{gray}{\# $\chi$ : B $\times$ D $\times$ N. N = H$*$W}
		\STATE
		\STATE \quad \textcolor{gray}{\#  generate $m$ part crops with crop scale $s$ for each sample.} 
		\STATE \quad  $C$ = crop($x$,$\;$$m$,$\;$$s$) \quad \textcolor{gray}{\#  $C$ : B $\times$ $m$ $\times$ 3 $\times $ h $\times$ w}
		 \STATE \quad  $C$ = $C$.reshape(B$*$$m$,$\;$3,$\;$h,$\;$w) \quad \textcolor{gray}{\#  $C$ : B$*m$ $\times$ 3 $\times$ h $\times$ w}
		 \STATE  
		 \STATE \quad  \textcolor{gray}{\#  map to feature space and obtain query vectors.} 
		 \STATE \quad $P$ = $avgpool(\mathcal{F}(C))$ \quad \textcolor{gray}{\#  $P$ : B$*$$m$ $\times$ D}
		 \STATE \quad  $P$ = $P$.reshape(B,$\;$ $m$,$\;$ D) \quad \textcolor{gray}{\#  $P$ : B $\times$ $m$ $\times$ D}
		 \STATE
		 \STATE \quad  \textcolor{gray}{\#  calculate similarity between queries $P$ and feature map $\chi$.} 
		 \STATE \quad $V$ = matmul($P$,$\;$ $\chi$) \quad \textcolor{gray}{\#  $V$ : B $\times$ $m$ $\times$ N} 
		 \STATE \quad \textcolor{gray}{\#  transfer to probability distribution}
		 \STATE \quad $V$ = Softmax($V$,$\;$ dim=2) 
		 \STATE
		 \STATE \quad \textcolor{gray}{\#  calculate uniform energy and gaussian energy}
		 \STATE \quad Generate two distribution $U$, $G$
		 \STATE \quad $E^u = KL\text{-}Div(V, U)$ \quad \textcolor{gray}{\#  $E^u$ : B $\times$ $m$}
		 \STATE \quad $E^g = KL\text{-}Div(V, G)$ \quad \textcolor{gray}{\#  $E^g$ : B $\times$ $m$}
		 \STATE
		 \STATE \quad \textcolor{gray}{\#  train an attacker} 
		 \STATE \quad Sort($E^u$, decending=True)  
		 \STATE \quad Sort($E^g$, decending=True) 
		 \STATE \quad $E$ = Concat($E^u$,$\;$ $E^g$, $\;$ dim=1) \quad \textcolor{gray}{\#  $E$ : B $\times$ $2*m$}
		 \STATE \quad logits = $f(E)$
		 \STATE \quad loss = $\mathcal{L}$(logits,$\;$ y)
		 \STATE \quad loss.backward()
		 \STATE \quad update($f$)
		 \RETURN $f$
	\end{algorithmic}
\end{algorithm}

The pseudo-code of PartCrop is in Algorithm~\ref{alg:Framwork}. Note that we do not specify where the training set and test set are from. In other words, we can use a public dataset to train an attacker with PartCrop and use this attacker to perform membership inference on other (even private) datasets. Though there usually exists distribution (\eg, domain) gaps between different datasets, our experiments in Sec~\ref{shadow} still show the effectiveness in such situation. We speculate the reason could be that the part-aware capability in self-supervised models is potentially less related to  distribution discrepancy. 

\section{Experimental Setting}

\subsection{Self-supervised Model Introduction}
We use three famous and highly cited self-supervised models that represent combinations of different self-supervised paradigms and structures as shown in Tab~\ref{tab:models}.  They are MAE~\cite{he2022masked}, DINO~\cite{caron2021emerging}, and MoCo~\cite{he2020momentum}. We detail them below. 

\textbf{MAE~\cite{he2022masked}} is one of representative \textit{masked image modeling} methods that predicts masked parts of an image from the remaining parts. MAE includes an encoder $\mathcal{E}$ and a decoder $\mathcal{D}$ that are constructed by Vision Transformer~\cite{DosovitskiyB0WZ21}. An image is first partitioned into patches, $ \rm{R} = \{\mathsf{R}_1, \mathsf{R}_2, \dots,
\mathsf{R}_N\}$. MAE randomly masks them with a given mask ratio (\eg, 75\% in the paper), partitioning $\rm{R}$ into two parts, visible (unmasked) patches $\rm{R}_{v}$ and masked patches $\rm{R}_{m}$. Then, the visible patches $\rm{R}_{v}$ are fed into encoder $\mathcal{E}$ to encode representation. The output representations along with mask tokens $\rm{T}$ that are randomly generated and learnable are served as input for decoder $\mathcal{D}$ to reconstruct the masked patches $\rm{R}_{m}$. The objective function is formulated as:
\begin{equation}
\mathcal{L} = MSE(\mathcal{D}(\mathcal{E}(\rm{R}_{v})\,, \rm{T})\,, \rm{R}_{m}) \,,
\end{equation}
where $MSE$ is the mean square error loss function. It measures the absolute distance between the reconstruction output of decoder $\mathcal{D}$ and the normalized RGB value of masked patches $\rm{R}_{m}$. After training, the encoder $\mathcal{E}$ is preserved for downstream tasks \eg, classification, while the decoder $\mathcal{D}$ is abandoned.
 
\textbf{DINO~\cite{caron2021emerging}} is one of representative \textit{contrastive learning} methods that maximizes the agreement of two augmented views from the same image. DINO contains two encoders named student $\mathcal{E}_{s}$ and teacher $\mathcal{E}_{t}$ that are also constructed by Vision Transformer~\cite{DosovitskiyB0WZ21}. Given an image, it is augmented with different strategies and derived into two views $v_{1}$ and $v_{2}$. Then, $v_{1}$ is fed into the student $\mathcal{E}_{s}$ followed by a projector $\mathcal{P}_{s}^{j}$ and a predictor $\mathcal{P}_{s}^{d}$ that consist of several \textbf{m}ulti-\textbf{l}ayer \textbf{p}erceptions (MLP). And $v_{2}$ is fed into the teacher $\mathcal{E}_{t}$ and a projector $\mathcal{P}_{t}^{j}$. The two output representations are going to agree with each other by the objective formulated as:
\begin{equation}
\mathcal{L} = CE\{\mathcal{P}_{s}^{d}(\mathcal{P}_{s}^{j}(\mathcal{E}_{s}(v_{1})))\,, \mathcal{P}_{t}^{j}(\mathcal{E}_{t}(v_{2}))\} \,,
\end{equation}
where $CE$ is cross-entropy loss function. By minimizing the loss, the two representations are brought close and become similar in representation space. After training, the student encoder $\mathcal{E}_{s}$ or the teacher encoder $\mathcal{E}_{t}$ is preserved for downstream tasks while the rest are abandoned. We use student encoder $\mathcal{E}_{s}$ in this work.  

\textbf{MoCo~\cite{he2020momentum}} is one of representative \textit{contrastive learning} methods in CNN era that brings two positive samples that are from the same image with different augmentation strategies close and pushes away negative samples from different images. MoCo consists of three important modules, two encoders, a vanilla encoder (denoted as $\mathcal{E}$) and a momentum encoder (denoted as $\mathcal{E}_{m}$), and a dynamic dictionary (denoted as $\Phi$). $\mathcal{E}$ and  $\mathcal{E}_{m}$ have the same structure, \eg, ResNet~\cite{resnet_he2016deep}, and output feature vectors. But $\mathcal{E}_{m}$ is updated slowly with \textbf{e}xponential \textbf{m}oving \textbf{a}verage (EMA) strategy. For clarity, we denote feature vectors from $\mathcal{E}$ as \textit{queries} $q$ and feature vectors from $\mathcal{E}_{m}$ as \textit{keys} $k$. $\Phi$ is a queue that pushes the output representations ($k$) from $\mathcal{E}_{m}$ in current mini-batch in queue and pops representations of early mini-batch. Given an image, MoCo firstly generates two different augmented samples $s_{1}$ and $s_{2}$. Then $s_{1}$ is fed into $\mathcal{E}$ and $s_{2}$ is fed into $\mathcal{E}_{m}$, producing two representations ($q_{s}$ and $k_{s}$) respectively. They are regarded as positive pairs as they are from the same image while $q_{s}$ and representations in $\Phi = \{\mathsf{k}_1, \mathsf{k}_2, \dots,
\mathsf{k}_N\}$ are negative pairs. In MoCo, InfoNCE loss~\cite{oord2018representation, he2020momentum, chen2020improved} is adopted as the objective function:
\begin{equation}
\mathcal{L} = -\log \frac{\exp(q_{s} \cdot k_{s} / \tau)}{\exp(q_{s} \cdot k_{s}/ \tau + \sum_{i=1}^{N}(q_{s} \cdot \mathsf{k}_i)/ \tau)} \,,
\end{equation}
where $\tau$ is the temperature coefficient. Optimizing such loss brings positive samples from the same image close while pushing away the negative samples in $\Phi$. In each iteration, $\Phi$ pops representations $\mathsf{k}$ of early mini-batch and pushes $k_{s}$ in queue as new $\mathsf{k}$. In this way, $\Phi$ is updated dynamically. After training, $\mathcal{E}$ is preserved for downstream tasks while the rest are discarded.

\setlength{\tabcolsep}{0.2cm}\begin{table}[h]
	\centering
	\small
	\begin{tabular}{c c c}
		\toprule
		\textbf{\textit{Model}} & 
		\textbf{\textit{Self-supervised Method}} &
		\textbf{\textit{Structure}}  \\
		\midrule
		\textit{MAE} & Masked Image Modeling & Vision Transformer \\
		\textit{DINO} & Contrastive Learning & Vision Transformer \\
		\textit{MoCo} & Contrastive Learning & CNN (ResNet) \\
		\bottomrule
	\end{tabular}
\caption{Different self-supervised models.}. 
	\label{tab:models}
\vskip -0.4in
\end{table}
\subsection{Dataset Introduction}

We conduct experiments on three widely used vision datasets in membership inference field including CIFAR10, CIFAR100, and Tinyimagenet. For each dataset, the training/test data partition setting follows the reference papers.

\textbf{CIFAR10 and CIFAR100}~\cite{cifar_10_krizhevsky2009learning} are two benchmark datasets. Both of them have $50,000$ training images and $10,000$ test images. CIFAR10 has $10$ categories while CIFAR100 has $100$ categories. The dimension for CIFAR10 and CIFAR100 images is $32 \times 32 \times 3$. During self-supervised pretraining, we only use their training set.

\textbf{Tinyimagenet}~\cite{tinyimagenet_le2015tiny} is another image dataset that contains $200$ categories. Each category includes $500$ training images and $50$ test images. The dimension for Tinyimagenet images is $64 \times 64 \times 3$. Similar to CIFAR10 and CIFAR100, we also only use its training set for self-supervised training.

\textbf{Dataset Split.} For simplicity, we follow previous work~\cite{Pruning_IJCAI, shejwalkar2021membership, zhu2022safety}, which assumes a strong adversary that knows 50\% of the (target) model's training data and 50\% of the test data (non-training data) to estimate attack performance's upper bound. As shown in Tab~\ref{data_split} the known data are used for training the attacker while the remaining datasets are adopted for evaluation. We also evaluate other smaller ratios with a sweep from 10\% to 50\% in Varying Adversary's Knowledge part of Appendix~\ref{app:more results}.

\textbf{Metric.} During evaluation, we report attack accuracy, precision, recall, and F1-score (F1) as measurement following EncoderMI~\cite{liu2021encodermi}. And we pay more attention to attack accuracy in our experiments as the accuracy intuitively reflects the performance of attack methods on distinguishing member and non-member and treat them equally while attack precision, recall, and F1 primarily focus on member data. Simultaneously, we observe that if an attack method produces an accuracy close to 50\%, the precision and recall often vary irregularly. In this circumstances, the derived F1 is meaningless. Therefore, accuracy may serve as a more appropriate metric for evaluating attack performance and determining whether this method fails to attack (reduces to random guessing).  

\setlength{\tabcolsep}{0.2cm}\begin{table}[h]
	\centering	
	\small
 \vskip -0.1in
	\begin{tabular}{l c  c | c  c}
		\toprule
		\multirow{2}{*}{\textbf{Datasets}}&
		\multicolumn{2}{c}{\textbf{Attack Training}}&\multicolumn{2}{c}{\textbf{Attack Evaluation}}\cr
		\cmidrule(lr){2-3} \cmidrule(lr){4-5}
		& $D^{known}_{train}$& $D^{known}_{test}$ &
		$D^{unknown}_{train}$ & $D^{unknown}_{test}$\\
		\midrule
		\textit{CIFAR10} & 25,000 & 5,000 & 25,000 & 5,000 \\
		\textit{CIFAR100} & 25,000 & 5,000 & 25,000 & 5,000 \\
		\textit{Tinyimagenet} & 50,000 & 5,000 & 50,000 & 5,000 \\ 
		\bottomrule
	\end{tabular}
\caption{Number of samples in dataset splits.}
	\label{data_split}
    \vskip -0.15in
\end{table}

\subsection{Baseline}\label{sec:baseline}
To show the effectiveness of PartCrop, three baselines are carefully considered. The first is a supervised model based attack method~\cite{ResAdv, zhu2022safety}. It is chosen to show that this attack method is not suitable for self-supervised models. For convenience, we denote this method as SupervisedMI. The second is label-only inference attack~\cite{choquette2021label}. It is similar to PartCrop and leverages a training-recipe-unrelated prior. The third is a recently proposed contrastive learning based attack method, \ie, EncoderMI~\cite{liu2021encodermi}. It is highly related to our research and mainly performs membership inference on contrastive learning model conditioned on knowing the training recipe.    

\textbf{SupervisedMI.} SupervisedMI mainly takes model output and label (one-hot form) as the input of attacker. Unfortunately, in self-supervised learning, label is absent. Hence, we repurpose this baseline by taking the model output as input to train attacker. Specifically, given an image $x$, we feed it into self-supervised model that outputs feature vector $\xi$. Then the feature vector $\xi$ is regarded as membership feature and fed into the attacker for training. Following previous works~\cite{ResAdv, zhu2022safety, liu2021ml}, we use a similar fully connected network as attacker. 

\textbf{Variance-onlyMI.} Generally, models are less sensitive to augmentations of training samples than of test samples. Label-only inference attack leverages this characteristic and take the collection of model-classified label from various augmented images as input for inference. However, in self-supervised learning, label is absent. To enable this attack, we repurpose this baseline by using various augmented images to compute the channel-wise variance in feature representations as attack input instead of classified labels. Intuitively, member data usually produce smaller variance.  We denote this modified attack as Variance-onlyMI. Following~\cite{choquette2021label}, we use the same augmentation strategies and a similar network as attacker.

\textbf{EncoderMI.} EncoderMI is the most related work with ours, which we regard as a strong baseline as it knows how the target model is trained in its setting. EncoderMI utilizes the prior that contrastive learning trained encoder is inclined to produce similar feature vectors for augmented views from the same image. Hence, it is necessary to know training recipe, especially the data augmentation strategies, to generate similar augmented views that appear in training. To construct this baseline, following~\cite{liu2021encodermi}, given an image $x$, EncoderMI creates $n$ augmented views using the same data augmentation strategies as self-supervised model. The $n$ augmented views are denoted as $x_{1}, x_{2}, \dots, x_{n}$. Then they are fed into self-supervised model to produce corresponding feature vectors ($\eta_{1}, \eta_{2}, \dots, \eta_{n}$). To obtain membership feature of $x$, EncoderMI calculates the similarity between $n$ feature vectors by the formula:
\begin{equation}
M(x) = \{S(\eta_{i}\,, \eta_{j}) \; | \; i\in [1\,,n], j\in [1\,,n], j > i \} \,,
\end{equation}
where $S(\,,)$ is the similarity function. In this research, we use cosine similarity and vector-based attacker setting as it performs the best among all the methods in ~\cite{liu2021encodermi}. When obtaining the membership feature, EncoderMI ranks the $n\cdot(n-1)/2$ similarity scores in descending order and feeds them into an attacker to train it. The attacker is also a fully connected neural network.  

\textbf{The discrepancy between EncoderMI and PartCrop.} In light of our previous introduction to EncoderMI, it is essential to delve into the discrepancy between EncoderMI and\textit{ PartCrop} to elucidate the novelty of this work.

$\bullet$ Firstly, in terms of granularity, EncoderMI operates at the image-level, whereas PartCrop operates at a finer-grained, part-level. Specifically, EncoderMI generates augmented images and computes similarities between these images (feature vectors), while PartCrop generates object parts (patches containing specific object parts) and calculates similarity distributions between each part (feature vector) and the entire image (feature map).

$\bullet$ Secondly, in terms of method, EncoderMI and PartCrop rely on different priors. EncoderMI utilizes the prior of contrastive learning, whereas PartCrop leverages the part-aware capability in self-supervised models.

$\bullet$ Thirdly, in terms of attack scope, unlike EncoderMI, which exclusively focuses on contrastive learning, PartCrop serves as a unified approach that can be employed in  contrastive learning, masked image modeling, as well as other paradigms, see Sec~\ref{unified}.

$\bullet$  Lastly, and perhaps most importantly, in terms of reality, we adopt a more realistic setting where adversaries are unaware of the training recipe (methods and training details), while EncoderMI assumes adversaries possess knowledge of the method (contrastive learning) and the associated details.

\textbf{Needed information assumed by each baseline.} The SupervisedMI and Label-onlyMI methods assume the victim model is supervised and necessitate model-classified labels to produce membership features. However, in the realm of self-supervised learning, labels are unavailable and the encoder only outputs image representations. Regarding EncoderMI, it is assumed to know training details and augmentation hyperparameters, which is unrealistic as self-supervised models often function as black-box services~\cite{Shaow_Learning} in practice. Consequently, these attacks on self-supervised models lack plausibility in reality. We also discuss the differences between PartCrop and general loss/entropy/confidence-based MI methods as well as the challenges of attacking self-supervised models compared to supervised models in Sec~\ref{sec:dis}.

\setlength{\tabcolsep}{0.15cm}{\begin{table*}[h]
		\begin{center}
			\small
   \vskip -0.05in
			\begin{tabular}{c| c | c c c c c c c c c c c c}
				\toprule
				\multirow{2}{*}{Method} & \multirow{2}{*}{Attacker} & \multicolumn{4}{c}{CIFAR100} & \multicolumn{4}{c}{CIFAR10}  & \multicolumn{4}{c}{Tinyimagenet} \cr
		\cmidrule(lr){3-6} \cmidrule(lr){7-10} \cmidrule(lr){11-14} 
                   & & Accuracy & Precision  &  Recall & F1 & Accuracy & Precision  &  Recall  & F1 & Accuracy & Precision  &  Recall  & F1 \\
				\midrule
				\midrule
				\multirow{4}{*}{\textit{MAE}} 
				 & \textit{SupervisedMI} & 50.62 & 50.95	& 33.32 & 40.29 & 50.32 & 50.43 &	36.80 & 42.55 & 50.00&50.00 &20.04 & 28.61  \\
                    & \textit{Variance-onlyMI} & 51.72 & 51.65	& 54.00	 & 52.80  & 51.52  & 51.60 &	49.16 & 50.35 & 50.10 & 50.10 & 50.82 & 50.46  \\
				 & \textit{EncoderMI} & 51.70  & 52.22 & 49.88 & 51.02  & 53.20 & 53.18 &	61.06 & 56.85 & 50.18	& 52.19	 & 18.32 & 27.12 \\
				 & \textit{PartCrop} &  \textbf{58.38}  & \textbf{57.73}	& \textbf{60.02} & \textbf{58.85} & \textbf{57.65}  & \textbf{57.60}	& \textbf{63.98} & \textbf{60.62} & \textbf{66.36}  &\textbf{62.68}	& \textbf{73.66}  & \textbf{67.73} \\
				\midrule 
				\multirow{4}{*}{\textit{DINO}}
				& \textit{SupervisedMI} & 49.60 &49.96&	12.04  & 19.40 & 50.50 & 50.52 & 47.82  & 49.13  & 50.34 &50.76&	0.66  & 1.30 \\
                    & \textit{Variance-onlyMI} & 50.71	& 50.55	&  	\textbf{64.82}	& \textbf{56.80}   & 58.82 & 58.48 & 60.80	 &  59.62 & 55.81 & 54.25 & \textbf{74.12} & \textbf{62.65}  \\
				& \textit{EncoderMI} & 55.52 &	57.20 &	44.82  &  50.26 & \textbf{66.40} & \textbf{65.55}& \textbf{63.60}  & \textbf{64.56} &  \textbf{63.87} & \textbf{66.73} &	54.96   & 60.27 \\
				& \textit{PartCrop} & \textbf{60.62}  & \textbf{66.91} & 	41.08  & 50.91 & 59.13 &  60.66	& 53.76  & 57.00 & 56.13  & 62.72 &31.60  & 42.02 \\
				\midrule
				\multirow{4}{*}{\textit{MoCo}}
				& \textit{SupervisedMI} & 49.94	& 49.42	& 5.16  & 9.34 & 50.31 &51.05&10.62  & 17.58 & 50.21  &  54.07&1.46  & 2.84 \\
                    & \textit{Variance-onlyMI} &  50.94  & 50.85 & 56.06  & 53.33  & 49.82  & 49.84 &	57.28 & 53.30 & 52.66 & 52.80 & 50.22 & 51.48  \\
				& \textit{EncoderMI} & 57.29 &	58.14 &	59.12  & 58.62 & 60.81 &	58.75 &	\textbf{72.24}  & 64.80 & 62.33 &	70.29&	41.26  &  51.99 \\
				& \textit{PartCrop} & \textbf{77.20} & \textbf{82.93} &	\textbf{67.84}   & \textbf{74.63} & \textbf{78.84}  & \textbf{86.46} & 66.94  & \textbf{75.46} &\textbf{73.77} & \textbf{76.02} & 	\textbf{69.20}  &  \textbf{72.45} \\
				\bottomrule
			\end{tabular}
		\end{center}
		\caption{Comparisions with SupervisedMI, Variance-onlyMI, and EncoderMI in partial setting.} 
		\label{tab:baseline}
  \vskip -0.25in
\end{table*}}

\subsection{Implementation Details}
We use official public code for self-supervised training~\footnote{MAE: https://github.com/facebookresearch/mae.}~\footnote{DINO: https://github.com/facebookresearch/dino.}~\footnote{MoCo: https://github.com/facebookresearch/moco}. Concretely, they are pretrained for 1600 epochs following their official setting including data augmentation strategies, learning rate, optimizers, \etc. Due to limited computational resources, we use Vision Transformer (small)~\cite{DosovitskiyB0WZ21} and set patch size to 8 for MAE and DINO. We use ResNet18~\cite{resnet_he2016deep} as backbone of MoCo.
We set batch size to 1024 for all models during pretraining. In PartCrop, for each image, we produce 128 patches that potentially contain parts of objects. Each patch is cropped with a scale randomly selected from 0.08-0.2. Afterward, we resize the part to 16 $\times$~16. When training attacker, following previous work~\cite{Shaow_Learning, ResAdv,zhu2022safety}, all the network weights are initialized with normal distribution, and all biases are set to 0 by default. The batch size is 100. We use the Adam optimizer with the learning rate of 0.001 and weight decay set to 0.0005. We follow previous work~\cite{liu2021ml, ResAdv} and train the attack models 100 epochs. In the training process, it is assured that every training batch contains the same number of member and non-member data samples, which aims to prevent attack model from being biased toward either side.

\section{Experiment}~\label{sec:experiment}
\subsection{Results in the Partial Setting}\label{partial}
To evaluate the effectiveness, we compare PartCrop, with SupervisedMI, Variance-onlyMI, and EncoderMI. We use the \textit{partial setting} where an adversary knows part of training and test data and report the results in Tab~\ref{tab:baseline}. 

It is observed that SupervisedMI achieves approximately 50\% attack accuracy across all scenarios and the precision and recall vary irregularly, suggesting its inability to attack self-supervised models trained on any of the datasets. 
It is reasonable as SupervisedMI is designed to attack supervised model. Generally, SupervisedMI should take supervised models' category probability distribution (in label space) as input while self-supervised models'output is encoded image feature (in representation space). This space gap leads to the failure of SupervisedMI.

Similarly, Variance-onlyMI fails across most scenarios while only performing well on DINO trained by CIFAR10 and Tinyimagenet. Also, its inference performance is inferior to EncoderMI and PartCrop. These results indicate that for Variance-onlyMI, utilizing variance as a characteristic for membership inference doesn't work for all SSL models.

EncoderMI performs well on MoCo and DINO, and even outperforms PartCrop in some cases. For example, EncoderMI outperforms PartCrop in Tinyimagenet by around 8\% on accuracy, 4\% on precision, 23\% on recall, and 18\% on F1. However, when performing membership inference on MAE, a masked image modeling based model, the attack accuracy of EncoderMI drops significantly (CIFAR100 and CIFAR10) and even reduces to random guess (Tinyimagenet). As for precision and recall, they decrease accordingly, especially for the recall on Tinyimagenet. The derived F1 is also lower than that of PartCrop. This contrast of results implies that EncoderMI heavily relies on the prior of training recipe of victim model to obtain satisfying performance. As mentioned before, EncoderMI is inspired by the characteristics that contrastive learning is inclined to produce similar feature vectors for augmented views from the same image. 
In the case of masked image modeling, which aims to reconstruct the masked portions, it is a generative method and differs significantly from contrastive learning, which is a discriminative method. Consequently, there does not exist such characteristic in masked image modeling and EncoderMI naturally fails to obtain satisfying results on MAE. Finally, we also consider adapting EncoderMI to attack MAE, \eg, replacing augmented views generated with patches. We conduct this experiment on CIFAR100. This adaptation produces 51.18 for accuracy, inferior to the original performance (51.70 Acc), further demonstrating the essential discrepancy between masked image modeling and contrastive learning.

In contrast, without knowing how self-supervised model is trained, PartCrop generally achieves satisfying attack performance on four metrics for all the models, which greatly demonstrates the effectiveness of our PartCrop. 

\subsection{Ablation Study}
To further understand PartCrop, we ablate three key elements in PartCrop including membership feature, crop number, and crop scale. The conducted experiments on CIFAR100 dataset below reveal how they influence the membership inference performance of our PartCrop, respectively. We primarily focus on the accuracy and also use F1 to represent the trade-off between precision and recall. 

\setlength{\tabcolsep}{0.5cm}{\begin{table}[h]
		\begin{center}
			\small
   \vskip -0.05in
			\begin{tabular}{c c c c c}
				\toprule
				Model & $E^{u}$ & $E^{g}$ & Accuracy & F1 \\ 
				\midrule
				\midrule
				\multirow{3}{*}{\textit{MAE}} 
				& \Checkmark  & & 57.43 &  54.98 \\
				&  & \Checkmark & 56.67 &	58.18 \\
				&  \Checkmark &  \Checkmark &  \textbf{58.38} &  \textbf{58.85} \\
				\midrule 
				\multirow{3}{*}{\textit{DINO}}
				& \Checkmark & & 54.16  &   48.39 \\
				&  & \Checkmark& 51.56 &  48.94 \\
				&\Checkmark &  \Checkmark &\textbf{ 60.62}  & \textbf{50.91} \\
				\midrule
				\multirow{3}{*}{\textit{MoCo}}
				& \Checkmark & & 70.27 &  68.97 \\
				&  & \Checkmark & 69.49 & 69.15 \\
				& \Checkmark & \Checkmark &\textbf{ 77.20} & \textbf{74.63} \\
				\bottomrule
			\end{tabular}
		\end{center}
		\caption{Ablation study on membership feature. $E^{u}$ is uniform energy and $E^{g}$ is gaussian energy. \Checkmark $\;$ means using the energy during training and evaluating attacker.} 
		\label{tab:membership feature}
		\vspace{-8pt}  
  \vskip -0.15in
\end{table}}

\subsubsection{Merbership Feature} In PartCrop, membership feature contains uniform energy $E^{u}$ and gaussian energy $E^{g}$. We investigate the impact of these factors separately and jointly on the attack performance. As shown in Tab~\ref{tab:membership feature},  $E^{u}$ and $E^{g}$ are both useful for membership inference. Moreover, we find that $E^{u}$ leads to higher attack accuracy than $E^{g}$ consistently among three models. For example, in MAE, $E^{u}$ produces 57.43\% for accuracy while $E^{g}$ is inferior and achieves 56.67\% accuracy. However, this situation is reversed when it comes to F1. When leveraging both $E^{u}$ and $E^{g}$, it further improves the attack performance on accuracy and F1.  This indicates that $E^{u}$ and $E^{g}$ are complementary to some extent. 

Besides, we also observe two interesting phenomena. The first is that MoCo (a CNN based model) is more vulnerable than MAE and DINO (two Vision transformer based models). PartCrop obtains about 70\%, even higher accuracy when attacking MoCo, outperforming the results on MAE and DINO by a large margin. Another is that DINO and MoCo (two contrastive learning based models) are more sensitive to the combination of $E^{u}$ and $E^{g}$ than MAE (masked image modeling based model) in terms of accuracy. Incorporating $E^{u}$ and $E^{g}$ leads to about 7\% improvement on attack accuracy and about 2\% and 5\% improvement on F1 for DINO and MoCo, greatly exceeding 2\% for MAE. Since exploring these phenomena is not our main goal in this research, we leave them as future work.

\subsubsection{Crop Number}
\begin{figure}	
\centering{\includegraphics[width=1\linewidth]{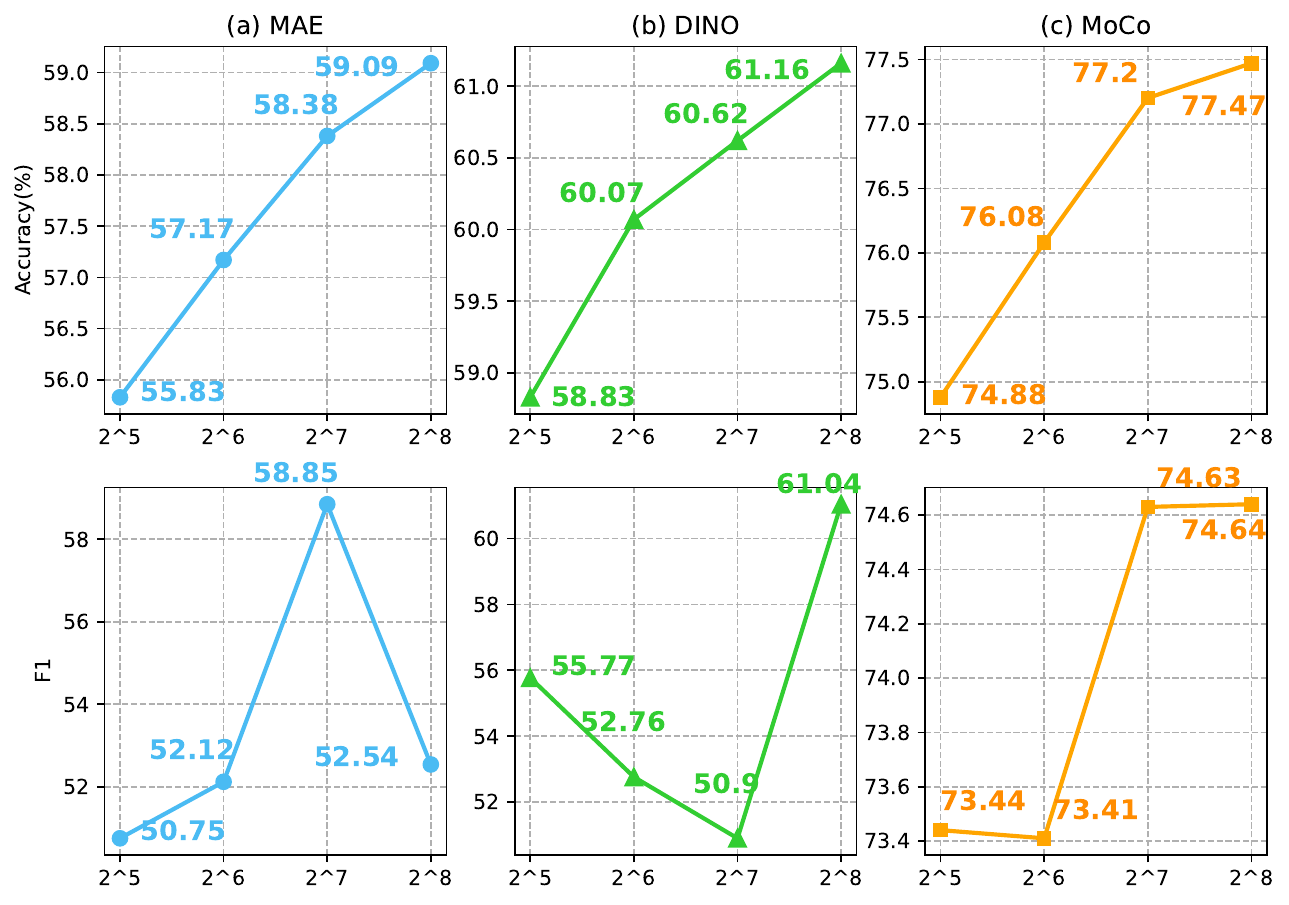}}
 \vskip -0.05in
	\caption{Ablation study on crop number. We consider four different crop number \ie, 32, 64, 128, and 256.}
	\label{fig:crop number}
 \vskip -0.2in
\end{figure}

The crop number in PartCrop is critical. It determines how many patches we can use. Intuitively, the more the crop number is, the more useful membership features are acquired, thereby leading to a higher attack performance. In this experiment, we consider four different crop numbers \ie, $32$, $64$, $128$, and $256$. Fig~\ref{fig:crop number} shows the relationship between query number and attack performance. Generally, more queries lead to better accuracy but more computation. Moreover, after a certain point, increasing crop numbers brings only marginal improvement. In terms of F1, we find that it exhibits opposite behavior, especially for MAE and DINO, indicating that crop number has different impact on F1, possibly because of the mutual competition between precision and recall. Considering that increasing crop number aggravates the computational burden, we select a trade-off (by using $128$ as our default setting) to ensure high attack performance while minimizing computation cost.

\begin{figure}	\centering{\includegraphics[width=1\linewidth]{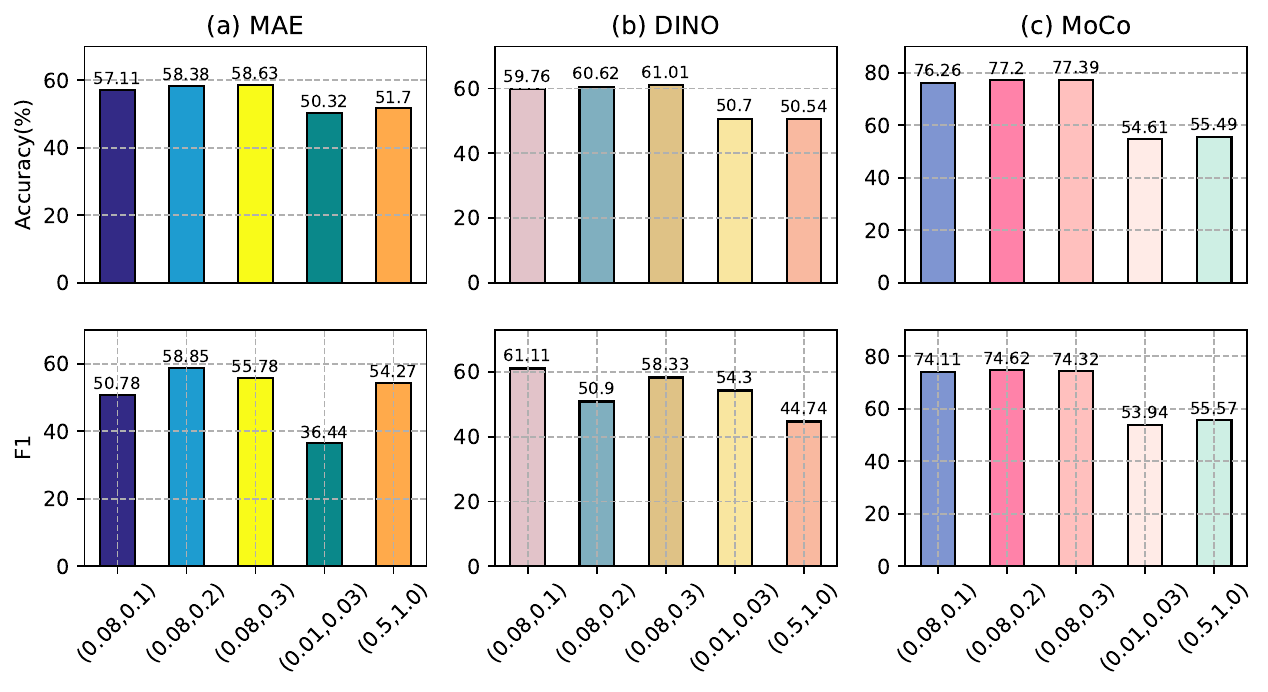}}
 \vskip -0.05in
	\caption{Ablation study on crop scale. We consider five different crop scales \ie, $(0.08,\; 0.1)$, $(0.08,\; 0.2)$, $(0.08,\; 0.3)$, $(0.01,\; 0.03)$, and $(0.5,\; 1.0)$.}
	\label{fig:crop scale}
 \vskip -0.15in
\end{figure}
\subsubsection{Crop Scale}\label{sec:crop_scale}
In PartCrop, crop scale determines how much ratio the area of the cropped patch is in that of raw image. We consider three crop scales setting including $(0.08,\; 0.1)$, $(0.08,\; 0.2)$, and $(0.08,\; 0.3)$. As presented in Fig~\ref{fig:crop scale}, it is shown that from $(0.08,\; 0.1)$ to $(0.08,\; 0.2)$, PartCrop produces about 1\% improvement for attack accuracy among three models.  As for F1, the influence is inconsistent: it significantly enhances the F1 of MAE; In contrast, it significantly reduces the F1 of DINO. Compared to DINO and MAE, MoCo undergoes mild F1 changes. We conjecture that this could be caused by the structure discrepancy: MAE and DINO are Vision Transformer while MoCo is CNN. 
While in scale $(0.08,\; 0.3)$, compared to $(0.08,\; 0.2)$, the attack accuracy is marginal, especially for MoCo. Hence, we use $(0.08,\; 0.2)$ as our default crop scale.

We also add two particular crop scales, \ie, $(0.01,\; 0.03)$ and $(0.5,\; 1.0)$,  to study the impact of extremely small and large crop scale. As shown in Fig~\ref{fig:crop scale}, $(0.01,\; 0.03)$ fails in membership inference for MAE and DINO and drops significantly for MoCo in terms of accuracy. It is reasonable as such scale makes the patches contain little useful part information. For another scale  $(0.5,\; 1.0)$, it causes the cropped image to cover most of the area of raw image, possibly including the whole object (in this situation, we can not call it part.). This violates our motivation of leveraging part capability in self-supervised models as the input is not a part of an object but more likely the whole. Consequently, the attack accuracy drops significantly. For F1, MAE obtains $36.44$ in $(0.01,\; 0.03)$ and DINO obtains $44.74$ in $(0.5,\; 1.0)$. As for MoCo, the F1 drops significantly under both $(0.01,\; 0.03)$ and $(0.5,\; 1.0)$.

\setlength{\tabcolsep}{0.08cm}{\begin{table*}[h]
		\begin{center}
			\small
			\begin{tabular}{c| c | c | c c c c c c c c c c c c c c c c}
				\toprule
				\multirow{2}{*}{Model} & \multirow{2}{*}{Pretrain Data}  & \multirow{2}{*}{Attack Data} & \multicolumn{4}{c}{SupervisedMI} & \multicolumn{4}{c}{Variance-onlyMI} & \multicolumn{4}{c}{EncoderMI}  & \multicolumn{4}{c}{PartCrop} \cr
		\cmidrule(lr){4-7} \cmidrule(lr){8-11} \cmidrule(lr){12-15} \cmidrule(lr){16-19} 
                &  &  & Acc & Pre  &  Rec & F1 & Acc & Pre  &  Rec & F1 & Acc & Pre  &  Rec & F1  & Acc & Pre  &  Rec  & F1 \\ 
				\midrule
				\midrule
				\multirow{6}{*}{\textit{MAE}} &
				\multirow{2}{*}{\textit{CIFAR100}} 
				& \textit{CIFAR10} &  50.29 & 50.65 & 22.46 & 31.12  & 49.78 & 49.75 & 44.38 & 46.91  & 50.05 & 55.81 & 0.48 & 0.95 & \textbf{56.54} & \textbf{57.48} &  \textbf{49.32} & \textbf{53.08} \\
				& & \textit{Tinyimagenet} &  50.39 & 72.41 & 1.26 & 2.48  & 50.05 & 50.02 & \textbf{93.34} & \textbf{65.13} & 50.21 & 5.30 &  52.06 & 9.62   & \textbf{55.63}  & \textbf{81.22} & 13.84 & 23.65 \\
				\cmidrule(lr){2-19}
				& \multirow{2}{*}{\textit{CIFAR10}} 
				& \textit{CIFAR100} & 50.00 & 50.00 & 0.22 & 0.43  & 50.19 & 50.77 & 12.56 & 20.14 & 49.92 &	49.95 &	\textbf{98.06} & \textbf{66.18} & \textbf{54.19} & \textbf{52.69} & 80.94  &  63.82 \\
				&& \textit{Tinyimagenet} &  50.25 & 50.71 & 17.82 & 26.37  & 49.98 & 49.99 & \textbf{66.00} & \textbf{56.89} & 49.91 & 39.53 & 0.34 & 0.67 & \textbf{52.13} & \textbf{99.51} & 4.14 &  7.95\\
				\cmidrule(lr){2-19}
				& \multirow{2}{*}{\textit{Tinyimagenet}}
				& \textit{CIFAR100} & 49.88 & 47.78 & 2.58  & 4.90  & 49.58 & 49.65 & 60.02 & 54.34 & 49.99 & 0.04 & 40.00 & 0.08 & \textbf{55.51} &	\textbf{54.71} & 	\textbf{64.06} & \textbf{59.02}  \\
				&& \textit{CIFAR10} & 49.99 & 46.15 & 0.12  &  0.24  &49.53 & 49.58 & 55.66 & 52.44& 49.62 & 2.58 & 43.58 & 4.87 & \textbf{54.46} &	\textbf{53.85}	& \textbf{62.42} & \textbf{57.82} \\
				\midrule 
				\multirow{6}{*}{\textit{DINO}} &
				\multirow{2}{*}{\textit{CIFAR100}} 
				& \textit{CIFAR10} & 49.49 & 46.44 & 6.66 & 11.65  & 51.73 & 55.16 & 18.50 & 27.71 &\textbf{ 57.54} & 54.81 & \textbf{79.78} & \textbf{64.98} & 55.22 & \textbf{56.97} & 	39.88 & 46.92\\
				& & \textit{Tinyimagenet} &49.86 & 49.90 & 71.20 & 58.67  & 50.20 & 50.44 & 23.20 & 31.78 & 50.00 & 50.00 & \textbf{100.00}  & \textbf{66.66} & \textbf{55.82} & \textbf{63.46} & 26.96 & 37.84 \\
				\cmidrule(lr){2-19}
				& \multirow{2}{*}{\textit{CIFAR10}} 
				& \textit{CIFAR100} & 50.02 & 50.10 & 9.72 & 16.28  & 50.27 & 50.18 & 75.08 & 60.16 & 49.99 & 50.00 & \textbf{99.90} & \textbf{66.64} & \textbf{54.86} & \textbf{69.69} & 15.82 & 25.78 \\
				&& \textit{Tinyimagenet} & 50.00 & 50.00 & \textbf{99.74} & \textbf{66.61}  & 53.87 & 56.82 & 32.24 & 41.14 & 50.11 & 50.09 & 98.80 & 66.48 &  \textbf{57.23} & \textbf{69.59}	& 25.68 & 37.52  \\
				\cmidrule(lr){2-19}
				& \multirow{2}{*}{\textit{Tinyimagenet}}
				& \textit{CIFAR100} & 50.02 & 50.01 & 98.54 & 66.34  &50.00 & 50.00 & 98.28 & 66.28 & 50.00 & 	50.00	& \textbf{100.00} & \textbf{66.66} & \textbf{52.38} & \textbf{51.72} &	80.60 & 63.00 \\
				&& \textit{CIFAR10} & 49.74 & 49.85 & 89.66 & 64.07  & 50.80 & 50.42 & 96.98 & 66.34 & 50.00 & 	50.00	& \textbf{100.00} & \textbf{66.66} & \textbf{54.00}  & \textbf{52.31}  & 	90.72  & 66.35 \\
				\midrule 
				\multirow{6}{*}{\textit{MoCo}} &
				\multirow{2}{*}{\textit{CIFAR100}} 
				& \textit{CIFAR10} & 49.98 & 49.97 & 30.56 & 37.93  & 49.30 & 49.27 & 47.06 & 48.14 & 54.38 & 52.43 & \textbf{94.48} & 67.44  &  \textbf{78.61}  & 	\textbf{78.31}	 &  79.14 & \textbf{78.72} \\
				& & \textit{Tinyimagenet} &  47.77 & 44.66 & 18.64 & 26.30  & 49.84 & 49.92 & 97.70 & 66.08 & 50.00  & 50.00  & \textbf{100.00} & 66.66  & \textbf{52.88} & \textbf{50.75} &	99.88  & \textbf{67.30} \\
				\cmidrule(lr){2-19}
				& \multirow{2}{*}{\textit{CIFAR10}} 
				& \textit{CIFAR100} & 50.04 & 50.81 & 2.50 & 4.77  & 49.87 & 49.86 & 46.60 & 48.17 & 55.88 & 63.54 & 27.58 & 38.46 & \textbf{74.32} &	\textbf{87.78} &	\textbf{56.50}  & \textbf{68.75} \\
				&& \textit{Tinyimagenet} & 49.84 & 49.88 & 64.58 & 56.29  & 49.58 & 49.77 & 93.06 & 64.85 & 51.44 & \textbf{92.35}	& 3.14 & 6.07 & \textbf{53.69} & 51.92 & \textbf{99.74} & \textbf{68.29} \\
				\cmidrule(lr){2-19}
				& \multirow{2}{*}{\textit{Tinyimagenet}}
				& \textit{CIFAR100} & 49.69 & 49.79 & 76.32 & 60.26  & 49.76 & 49.87 & \textbf{93.84} & \textbf{65.13} & 50.77 & 64.98 & 3.34 &  6.35 &\textbf{56.48} & \textbf{100.00} & 12.96 &  22.95 \\
				&& \textit{CIFAR10} & 49.93 & 49.74 & 13.74 & 21.53  & 49.31 & 49.63 & \textbf{93.08} & \textbf{64.74} & 52.09 & 64.30 & 9.40 & 16.40 & \textbf{62.94} & \textbf{99.84} & 25.92 & 41.16 \\
				\bottomrule
			\end{tabular}
		\end{center}
		\caption{Compare PartCrop with three baselines in shadow setting. Acc: Accuracy, Pre: Precision, Rec: Recall} 
		\label{tab:shadow}
		\vspace{-4pt}  
  \vskip -0.2in
\end{table*}}

\subsection{Results in the Shadow Setting} \label{shadow}
Besides the partial setting, we also conduct experiments in a \textit{shadow setting} where an adversary has no knowledge about the training data and test data, but attacks it via training an attacker on a public dataset. Specifically, for simplicity, we use the data split from Tab~\ref{data_split}, \eg, 50\% known data of CIFAR100, to train an attacker, and then attack a victim encoder trained by TinyimageNet (or CIFAR10), inferring their 50\% unknown data.

As shown in Tab~\ref{tab:shadow}, we perform cross-dataset membership inference for all the models and compare the results with SupervisedMI, Variance-onlyMI, and EncoderMI. We see that SupervisedMI produces around 50\% for accuracy, failing in all situations. For Variance-onlyMI, it fails to infer effectively in most cases where approximately 50\% accuracy is achieved.  For EncoderMI, similar to \textit{partial setting}, it fails to work on MAE. On DINO, only EncoderMI trained on CIFAR100 successfully attacks CIFAR10 while the rest all fail. On MoCo, there are still some failure cases but the whole performance is improved compared to DINO. 
In contrast, though PartCrop achieves inferior performance compared with EncoderMI under certain conditions, PartCrop succeeds in all cross-dataset attacks and produce most of the best results in various cross-dataset settings. This strongly demonstrates the generalization of PartCrop to datasets with distribution discrepancy compared to EncoderMI.

\subsection{Potential Tapping}
In this section, we aim to further explore PartCrop's potential. First, we consider the impacts of stronger data augmentations on PartCrop, as these are potential improvements for self-supervised models. Next, we verify the generalization of PartCrop to more self-supervised models and datasets. We also evaluate PartCrop across more self-supervised paradigms to demonstrate its broad effectiveness. Additionally, we illustrate PartCrop's efficacy with less adversary knowledge, \ie, the proportion of the known part of a target dataset, show PartCrop's complementarity to other methods, demonstrate its effectiveness on image datasets from other domains, and evaluate its inference runtime. The experimental results are provided in Appendix~\ref{app:more results} due to space limitations.

\setlength{\tabcolsep}{0.15cm}{\begin{table*}[h]
		\begin{center}
			\small
   \vskip -0.05in
			\begin{tabular}{c| c c c c c c c c c c c c c c c c}
				\toprule
    \multirow{2}{*}{Attacker} & \multicolumn{4}{c}{MoCo} & \multicolumn{4}{c}{Random Rotataion}  & \multicolumn{4}{c}{Mix-up} & \multicolumn{4}{c}{Adversarial learning} \cr
		\cmidrule(lr){2-5} \cmidrule(lr){6-9} \cmidrule(lr){10-13} \cmidrule(lr){14-17}
                   & Acc & Pre  &  Rec  & F1 & Acc & Pre  &  Rec & F1 &  Acc & Pre  &  Rec & F1 & Acc & Pre  &  Rec & F1 \\
				\midrule
				\midrule
				 \textit{SupervisedMI} & 49.94	& 49.42	& 5.16 & 9.34 & 49.98 & 49.98 & 60.66 &  54.80 & 50.18 & 50.74 &  12.30 & 19.80 & 50.36 & 50.49 & 6.14 & 10.94  \\
                    \textit{Variance-onlyMI} & 50.94  & 50.85 & 56.06  & 53.33  & 51.01 & 51.47 & 35.22 & 41.82  & 50.20 & 50.20 & 50.12 & 50.16 & 50.07  & 50.12 & 30.56 & 37.97   \\
				 \textit{EncoderMI} & 57.29 &	58.14 &	59.12 & 58.62	 & 54.92 &	56.21	& 44.52 &  49.68 & 52.82 &\textbf{56.96} &	23.08 & 32.8 & 51.80 & 	51.88 &	49.60 & 50.71  \\
				 \textit{PartCrop} & \textbf{77.20} & \textbf{82.93} &	\textbf{67.84} & 	\textbf{74.63}  &   \textbf{73.85} &	\textbf{81.66} &	\textbf{61.52}  & \textbf{70.17} & \textbf{56.94} &	55.35 & \textbf{71.76} & \textbf{62.50}& \textbf{68.93} &	\textbf{65.98} & \textbf{78.16} & \textbf{71.56} 	\\
				\bottomrule
			\end{tabular}
		\end{center}
		\caption{Compare PartCrop with three baselines on extra data augmentation. Acc: Accuracy, Pre: Precision, Rec: Recall} 
		\label{tab:data aug}
  \vskip -0.2in
\end{table*}}

\begin{figure}
\centering{\includegraphics[width=1\linewidth]{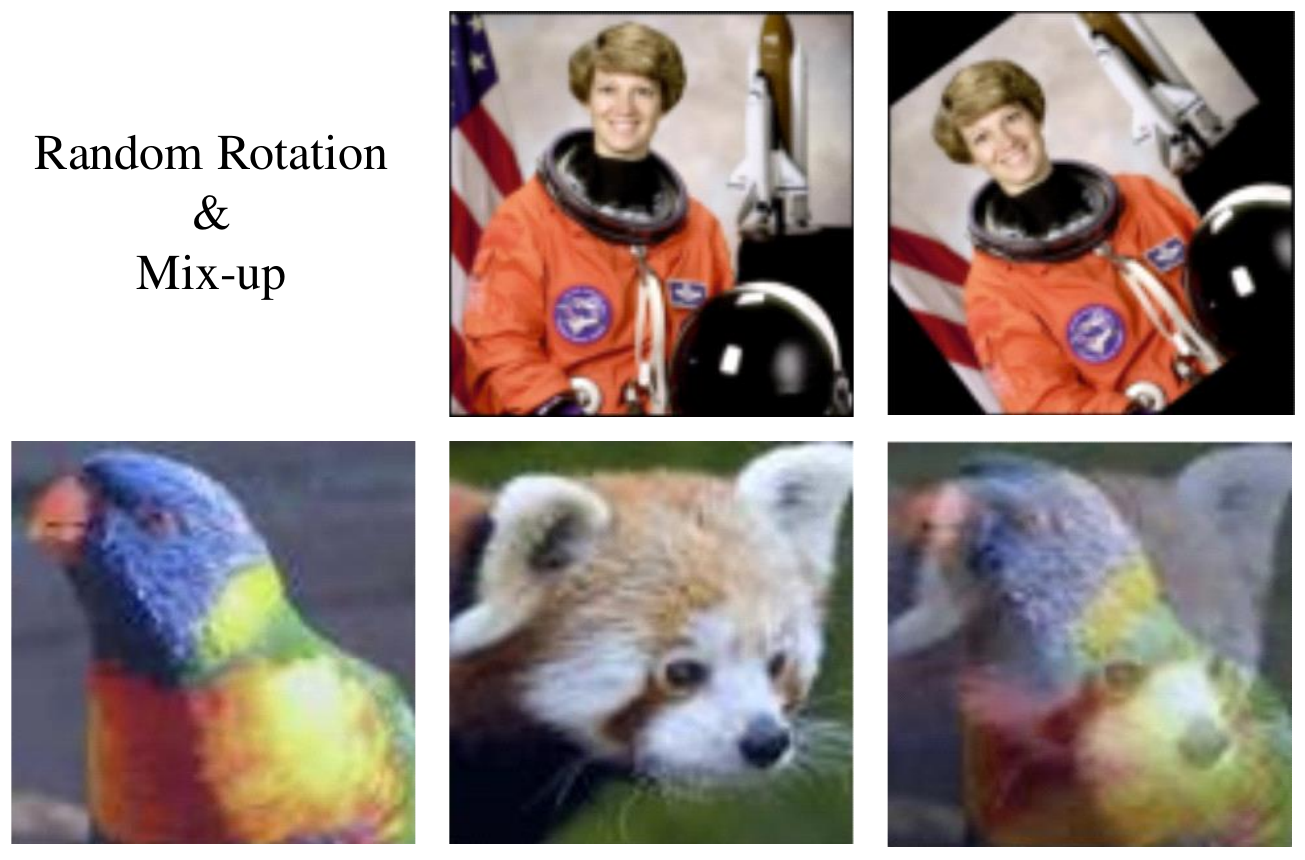}}
	\caption{Illustration of random rotation and mix-up. The first row is random rotation. The second row is mix-up~\cite{zhang2018mixup}.}
	\label{fig:Illustration}
	\vspace{-8pt}
 \vskip -0.05in
\end{figure}

\subsubsection{Impacts of Data Augmentation}
Data augmentation plays a vital role in self-supervised learning. Although Tab~\ref{tab:baseline} demonstrates the favorable performance of PartCrop under various data augmentation strategies employed in the three models, we aim to introduce additional stronger data augmentations to further validate the superiority of PartCrop in comparison to SupervisedMI, Variance-onlyMI, and EncoderMI. Specifically, we explore three types of data augmentations including random image rotation [-180$^{\circ}$, 180$^{\circ}$] (intra-image operation), mix-up~\cite{zhang2018mixup}~\footnote{Mix-up obtains a combination of two images with a coefficient. It is used in contrastive learning~\cite{kim2020mixco}. Code is at https://github.com/Lee-Gihun/MixCo-Mixup-Contrast} (inter-image operation), and adversarial learning~\cite{szegedy2014intriguing, biggio2013evasion}~\footnote{Adversarial learning is also used in contrastive learning~\cite{kim2020adversarial}. Code is at https://github.com/Kim-Minseon/RoCL} (data training manner). In~\cite{kim2020mixco} and~\cite{kim2020adversarial}, we find that both methods are built upon MoCo. Hence, we adopt MoCo as the baseline and utilize their official code to train the self-supervised models on CIFAR100. Regarding random image rotation ([-180$^{\circ}$, 180$^{\circ}$]), we incorporate it into the original data augmentation recipe of MoCo. The results of our attacks are presented in Tab~\ref{tab:data aug}. The introduction of extra data augmentation has minimal impact on SupervisedMI since it essentially relies on random guessing. In contrast, these data augmentations generally lead to a certain degree of decrease in attack accuracy for both EncoderMI, Variance-onlyMI~\footnote{We notice a slight improvement in Variance-onlyMI when adding random rotation, which is because Variance-onlyMI follows Label-onlyMI~\cite{choquette2021label} to adopts random rotation, and unintendedly mimics target model to augment images to generate membership feature, which could potentially resemble that of augmented training images.}, and PartCrop. This outcome is reasonable as these data augmentations mitigate overfitting to the training set. Fig~\ref{fig:Illustration} demonstrates that random rotation can result in the loss of some image information, which is replaced with black regions in the four corners. Mix-up combines two images and introduces blurring, thereby impairing the clear perception of the content of each image by self-supervised models and hindering the part-aware capability. The accuracy of PartCrop notably decreases in the case of Mix-up comapred to other data augmentations. Adversarial learning adopts an adversarial manner to prevent overfitting in self-supervised models. Compared to PartCrop, EncoderMI is more susceptible to the effect of adversarial learning. However, it is seen that PartCrop consistently outperforms EncoderMI across all data augmentations, indicating the superiority of PartCrop in comparison to SupervisedMI, Variance-onlyMI, and EncoderMI.


\subsubsection{Model and Dataset Generalization} As PartCrop is superior over EncoderMI, especially on masked image modeling methods, to further validate the effectiveness, we conduct experiments on another two models, \ie, CAE~\cite{chen2022context} (MIM) and iBOT~\cite{zhou2021ibot} (combine CL and MIM) using CIFAR100. As shown in Tab~\ref{tab:generalization}, In CAE, PartCrop (58.27\% Acc) outperforms EncoderMI (49.97\% Acc) by 8\%. In iBOT, PartCrop (60.03\% Acc) outperforms EncoderMI (51.73\% Acc) by 8\%. These results further verifies the generalization of PartCrop on models over EncoderMI.

On the other hand, we note that the evaluation datasets used currently are mostly single-object images. To verify the effectiveness of PartCrop on multiple-object image datasets, \eg, MS COCO~\cite{lin2014microsoft}, we conduct attack on Long-seq-MAE~\cite{hu2022exploring} pretrained on COCO. In Tab~\ref{tab:generalization}, PartCrop obtains 55.15\% Acc while EncoderMI (50.13\%) completely fails. This is reasonable as EncoderMI is designed for CL, \eg, MoCo, which often trains on single-object images. It is worth noting that this type of experiment is absent in EncoderMI and by contrasting, we show the generalization of PartCrop on datasets as well as potentially more significant value in reality application.

\setlength{\tabcolsep}{0.10cm}{\begin{table}[h]	
  \begin{center}
			\small
			\begin{tabular}{c| c | c c c c}
				\toprule
				\multirow{2}{*}{Method} & \multirow{2}{*}{Attacker} & \multicolumn{4}{c}{CIFAR100}\cr
		\cmidrule(lr){3-6} 
                   & & Accuracy & Precision  &  Recall & F1 \\
				\midrule
				\midrule
				\multirow{4}{*}{\textit{CAE}} 
				 & \textit{SupervisedMI} & 50.24  & 50.25	& 47.60 & 48.89 \\
                    & \textit{Variance-onlyMI} & 50.04  & 	50.03 & 62.40  & 55.53 \\
				 & \textit{EncoderMI} &  49.97 & 49.98 & \textbf{85.18} & \textbf{63.00}  \\
				 & \textit{PartCrop} &  \textbf{58.27}  & \textbf{59.53}	& 51.66 & 55.32 \\
				\midrule 
				\multirow{4}{*}{\textit{iBOT}}
				& \textit{SupervisedMI} &  49.94	& 42.50	&  0.34 & 0.67  \\
    & \textit{Variance-onlyMI} & 49.44 & 49.57	& \textbf{64.30} & 55.98  \\
				& \textit{EncoderMI} & 51.73 & 52.25	 &	40.04  & 45.34  \\
				& \textit{PartCrop} & \textbf{60.03} & \textbf{60.17} & 	59.34  & \textbf{59.75} \\
				\midrule
				\multirow{4}{*}{\makecell[c]{Long$-$seq MAE \\ (MS COCO)}}
				& \textit{SupervisedMI}  &  49.17 &  49.26 &	55.82 & 52.34  \\
    & \textit{Variance-onlyMI} & 50.26 & 50.41 & 31.92 & 39.09 \\
				& \textit{EncoderMI} & 50.13  & 50.65	& 10.16 & 16.92  \\
				& \textit{PartCrop} & \textbf{55.15} & \textbf{54.99} &	\textbf{56.80}  & \textbf{55.88} \\
				\bottomrule
			\end{tabular}
		\end{center}
		\caption{Results on extra self-supervised models.} 
		\label{tab:generalization}
		\vspace{-8pt}  
  \vskip -0.15in
\end{table}}

\setlength{\tabcolsep}{0.15cm}{\begin{table}[h]		
  \begin{center}
			\small
			\begin{tabular}{c| c | c c c c c c c c c c c c}
				\toprule
				\multirow{2}{*}{Method} & \multirow{2}{*}{Attacker} & \multicolumn{4}{c}{CIFAR100}\cr
		\cmidrule(lr){3-6} 
                   & & Accuracy & Precision  &  Recall & F1 \\
				\midrule
				\midrule
				\multirow{4}{*}{\textit{SWaV}} 
				 & \textit{SupervisedMI} & 49.97  & 49.88	& 12.78 & 20.35  \\
     & \textit{Variance-onlyMI} & 50.22  & 50.19 	& 59.12 & 54.29 \\
				 & \textit{EncoderMI} & 53.50  & 53.79 & 49.68 & 51.65  \\
				 & \textit{PartCrop} &  \textbf{68.98}  & \textbf{69.06}	& \textbf{68.76} & \textbf{68.91} \\
				\midrule 
				\multirow{4}{*}{\textit{VICReg}}
				& \textit{SupervisedMI} & 49.38 & 49.42 & 53.00 & 51.14 \\
    & \textit{Variance-onlyMI} &  49.79 & 49.83	& \textbf{59.88} & \textbf{54.39}  \\
				& \textit{EncoderMI} & 50.56 & 50.95 & 50.55 & 50.75  \\
				& \textit{PartCrop} & \textbf{55.31} & \textbf{80.55} & 14.00 & 23.85 \\
				\midrule
				\multirow{4}{*}{\textit{SAIM}}
				& \textit{SupervisedMI} & 50.30	&  50.24 & 61.58 & 55.33 \\
    & \textit{Variance-onlyMI} & 50.35 & 50.42 & 42.38 & 46.05 \\
				& \textit{EncoderMI} & 50.72 & 50.55 & 65.66 & 57.12 \\
				& \textit{PartCrop} & \textbf{56.74} & \textbf{55.60} & \textbf{66.86}  & \textbf{60.71} \\
				\bottomrule
			\end{tabular}
		\end{center}
		\caption{Results on other self-supervised paradigms.} 
		\label{tab:unified}
		\vspace{-8pt} 
  \vskip -0.2in
\end{table}}

\subsubsection{Evaluation on Other Paradigms} \label{unified}
Besides contrastive learning and masked image modeling, we additionally consider three representative (or recently proposed) models trained by extra self-supervised paradigms including clustering method (SwAV~\cite{caron2020unsupervised}), information maximization (VICReg~\cite{bardes2021vicreg}), and auto-regressive methods (SAIM~\cite{qi2023exploring}), and conduct experiments on CIFAR100. As reported in Tab~\ref{tab:unified}, PartCrop obtains 68.98\% Acc for SwAV, 55.31\% Acc for VICReg, and 56.74\% Acc for SAIM, outperforming EncoderMI (53.50\%, 50.56\%, and 50.72\%). For the attack success, we conjecture that these models may possess characteristics resembling part-aware capability. Studies in~\cite{chen2022intra} empirically proves that the representation space tends to preserve more equivariance and locality (part), supporting our speculation. On the other hand, these results also indicate that PartCrop is a more unified method than EncoderMI in the field of visual self-supervised learning.

\section{Defenses}
Although our primary focus is to propose a new attack method, there are some commonly used defense methods for MI, \ie, early stop and differential privacy. Hence, for the integrity of our work, we follow the existing research~\cite{liu2021encodermi, jia2022badencoder, liu2022poisonedencoder, liu2022stolenencoder} and conduct an experimental verification on whether these methods are effective against our newly proposed method. We perform inference attack and classification task~\footnote{We use pretrained model to extract feature and feed it into a learnable linear classification layer following~\cite{liu2021encodermi}.} on CIFAR100. In our experiments, we find that they are not very effective, thus We newly introduce a defense approach called shrinking crop scale range that can better defend against PartCorp. We organize all results in Tab~\ref{tab:all} for better comparison.

\textbf{Early Stop (ES).}$\;$ PartCrop is inspired by the characteristic that the training data is more discriminative than test data in part (stronger part response). This is mainly caused by overfitting. A direct and effective way to avoid overfitting is early stop that trains model with less training time. We train MAE and MoCo with 800 epochs compared to vanilla (1600 epochs) and train DINO for 400 compared to vanilla (800 epochs)~\footnote{We do not train DINO 1600 epochs in this section as it makes DINO overfitting on training set, which leads to inferior task performance on test set. Specifically, 800 epochs produce 57.1\% classification accuracy while 1600 epochs produce 47.0\%.}. It is shown in Tab~\ref{tab:all} that ES is generally effective to decrease the membership 
inference accuracy for all models. 
As a price, ES reduces the classification accuracy by a large margin for all models.

\textbf{Differential Privacy (DP).}$\;$ Differential privacy is widely used to preserve membership privacy due to theoretical guarantees. Concretely, it could add noise to training data~\cite{duchi2013local}, objective function~\cite{iyengar2019towards, jayaraman2019evaluating}, and backpropagation gradient~\cite{dp_abadi2016, shokri2015privacy, jayaraman2019evaluating}. For example, DP-SGD~\cite{dp_abadi2016} adds random Gaussian noise to the gradient calculated by stochastic gradient descent. We use Opacus and follow the official hyperparameters~\footnote{Please see \url{https://github.com/pytorch/opacus/blob/main/tutorials/building_image_classifier.ipynb}} to implement our DP training. As shown in Tab~\ref{tab:all}, DP enhances defense capability of self-supervised models, especially for MoCo.
Specifically, the attack accuracy and precision both decrease for all the models. In particular, MoCo has degenerated to random guess. On the other hand, differential privacy also incurs significant utility loss, especially for DINO. Hence, more effective methods are expected and thereby we propose SCSR.

\setlength{\tabcolsep}{0.25cm}{\begin{table}[h]
		\begin{center}
			\small
   \vskip -0.05in
			\begin{tabular}{c| c | c c c c | c }
				\toprule
				\multirow{2}{*}{Model} & \multirow{2}{*}{Defense} & \multicolumn{4}{c}{Attack} & Task \cr
		\cmidrule(lr){3-6} \cmidrule(lr){7-7} 
                   & & Acc ($\downarrow$) & Pre  &  Rec & F1 & Acc($\uparrow$) \\
				\midrule
				\midrule
				\multirow{4}{*}{\textit{MAE}} 
				 & \textit{Vanilla} & 58.4 &   57.7  & 60.0 & 58.8 & 48.3 \\
				 & \textit{ES} & 56.9 & 60.3  & \underline{44.3} & \underline{51.1}  & \underline{45.1} \\
				 & \textit{DP} & \underline{54.5} &   \textbf{53.4}  & 80.0 & 64.0 & 13.2 \\
        & \textit{SCSR} &  \textbf{54.3} & \underline{59.6}  & \textbf{39.2} & \textbf{47.3} & \textbf{46.8} \\	
    \midrule 
				\multirow{4}{*}{\textit{DINO}} 
				 & \textit{Vanilla} & 58.1 & 59.4  &  48.3 & 53.3  & 57.1 \\
				 & \textit{ES} & 57.0 & 58.5 & \textbf{51.1} & \textbf{54.5} & \underline{49.8} \\
				 & \textit{DP} &  \textbf{54.4} & \textbf{53.3}  &  \underline{71.4} & \underline{61.0}  & 11.7 \\
        & \textit{SCSR} &  \underline{55.1} & \underline{53.6} &  74.5  & 62.3 & \textbf{50.9} \\
				\midrule
				\multirow{4}{*}{\textit{MoCo}} 
				 & \textit{Vanilla} & 77.2 &   82.9   & 67.8  & 74.6 & 44.0 \\
				 & \textit{ES} & 74.6 & 82.9   & \textbf{62.7} &  71.4 & \textbf{41.4} \\
				 & \textit{DP} & \textbf{50.0} &   \textbf{50.0}   & 100.0  & \textbf{63.9} & 23.1 \\
        & \textit{SCSR} &  \underline{71.3} & \underline{71.1}   & \underline{68.2}  &  \underline{69.6} & \underline{35.4} \\
				\bottomrule
			\end{tabular}
		\end{center}
		\caption{Results on defense via three methods including early stop, differential privacy, and shrinking crop scale range in self-supervised learning. Bold (\textbf{54.3}) is the best result and underline \underline{55.1} indicates the second-best result.} 
		\label{tab:all}
  \vskip -0.2in
\end{table}}

\textbf{Shrinking Crop Scale Range (SCSR).}$\;$
In self-supervised learning, random crop is a crucial data augmentation strategy, especially for contrastive learning. Generally, self-supervised models are trained with a wide range of crop scale, \eg, $(0.2,\; 1)$. Our insight is that this potentially contributes to self-supervised models' part-aware capability, which is used in PartCrop. Thus, a simple idea is to shrink crop scale range by increasing the lower bound. For example, we can use $(0.5,\; 1)$ instead of vanilla setting ($(0.2,\; 1)$). In this way, self-supervised models fail to see small patches. In Tab~\ref{tab:all} the attack accuracy decreases for all the models. Moreover, SCSR brings significant drop to MAE in recall and to DINO and MoCo in precision. Naturally, we also observe a task accuracy drop on them. 
When comparing the SCSR with ES and DP, we observe that the newly proposed method exhibits greater effectiveness. SCSR produces most of the best and second-best results in terms of the privacy and utility. However, truthfully, the absolute effect is only somewhat satisfactory, particularly for MoCo. Therefore, further research is required to explore more effective defense methods against PartCrop in the future.

\textbf{A two-stage strategy to select the lower bound of SCSR.} The lower bound is a key hyperparameter for defending against PartCrop. Hence, we can leverage a two-stage  coarse-to-fine strategy. In the first stage, we choose a wide initial lower bound range, \eg, 0.3, 0.4, and 0.5, and assess the defense capability for each configuration. Then, in the second stage, we further select the two best-performing lower bounds and proceed with finer adjustments between them using smaller increments, \eg, 0.02. In this way, we can fine-tune the lower bound with greater precision, ensuring optimal defense against PartCrop.


\section{Related Work}
\textbf{Membership Inference:} Membership inference~\cite{Shaow_Learning, ResAdv} aims to infer whether a data record is used for training a deep neural model. Usually, a neural model exhibits different behavior on training data and test data as prior arts~\cite{carlini2019secret, nasr2019comprehensive} find that neural models are easily overfitting training set. Neural models often produce high confidence for training data in classification. Based on this observation, researchers propose numerous membership inference methods, \eg, binary classifier based methods~\cite{Shaow_Learning, ResAdv} and metric based methods~\cite{dp_yeom2018privacy, pre_conf_ML_Leaks, song2021systematic_entropy}. Besides classification, membership inference has been widely extended to other fields~\cite{hayes2017logan, song2020information, mahloujifar2021membership, duddu2020quantifying, tseng2021membership, duan2024membership}. For example, MI is investigated on generative models~\cite{hayes2017logan, goodfellow2020generative}. Song and Raghunathan~\cite{song2020information} introduced the first MI on word embedding and sentence embedding models that map raw objects (\eg, words, sentences, and graphs) to real-valued vectors with the aim of capturing and preserving important semantics about the underlying objects.  Duddu~\etal ~\cite{duddu2020quantifying} introduce the first MIA on graph embedding models.  M$^4$I~\cite{hu2022m} is the first to investigate membership inference on multi-modal models. Liu~\etal propose the first membership inference approach called EncoderMI~\cite{liu2021encodermi} for contrastive learning based models. Different from EncoderMI, we go further and conduct membership inference on self-supervised models without training recipe in hand.  

\textbf{Membership Inference Defense:} Considering that membership inference has the potential to raise severe privacy risks to individuals, studying how to defend against it becomes important. Differential privacy~\cite{dp_abadi2016, dp_rahman2018} (DP) is proposed to disturb the learning of neural attackers by adding noise to training data or models. Memguard~\cite{memguard_jia2019} is proposed to add carefully designed noise to the output of target models. Nasr~\etal leverage adversarial regularization~\cite{ResAdv} by involving an attacker during training. Shejwalkar and Houmansadr~\cite{shejwalkar2021membership} attempt to leverage knowledge distillation~\cite{KD_hinton2015distilling} (KD) to help improve target model defense ability. Different from teacher-student KD mode, Tang~\etal propose SELENA~\cite{tang2022mitigating} which contains two major components including a novel ensemble architecture for training called Split-AI and a self-distillation strategy. More recently, model compression technicals have been used in defending against MI. MIA-Pruning~\cite{Pruning_IJCAI} prunes model size to avoid target model being overfitting to training set. But Yuan \& Zhang~\cite{pruning_defeat_yuan2022membership} find that pruning makes the divergence of prediction confidence and prediction sensitivity increase and vary widely among different classes of training (member) and test (no-member) data. Zhu~\etal propose SafeCompress~\cite{zhu2022safety, zhu2024safety} to bi-optimize MI defense capability and task performance simultaneously when compressing models. 

\textbf{Self-Supervised Learning:} Current self-supervised methods can be categorized into different families among which masked image modeling~\cite{bao2021beit, he2022masked, chen2022context, zhou2021ibot} and contrastive learning~\cite{chen2020simple, chen2020improved, chen2021empirical, caron2021emerging, grill2020bootstrap} are the most representative and prevalent. Masked image modeling aims to reconstruct masked part of an image via an encoder-decoder structure.  For example, MAE~\cite{he2022masked} directly reconstructs the masked RGB color and further boosts the downstream performance. For contrastive learning, MoCo~\cite{he2020momentum}, as one of represetatives, proposes a momentum strategy. Further, in the transformer era, DINO~\cite{caron2021emerging} explores new properties derived from self-supervised ViT and accordingly designs a learning strategy interpreted as a form of self-distillation with no labels. There are also works trying to understand self-supervised methods~\cite{xie2022revealing, kong2022understanding, zhu2023understanding, saunshi2022understanding, chen2022intra, zhong2022self, wei2022contrastive}. Zhu~\etal~\cite{zhu2023understanding} show that masked image modeling and contrastive learning both have superior part-aware capability compared to supervised learning. Besides boosting and understanding self-supervised models, due to its amazing potential in leveraging massive unlabelled data, studying how to attack and protect self-supervised models is also critical~\cite{jia2022badencoder, liu2022poisonedencoder, liu2022stolenencoder, saha2022backdoor, dziedzic2022dataset, li2022demystifying, cong2022sslguard, liu2021encodermi}. 


\section{Discussion}\label{sec:dis}

\textbf{Differences from the general MI methods based on model loss/entropy/confidence.} 
Though general MI methods and PartCrop both focus on inferring a model's member data, they leverage different characteristics of the model.
Specifically, general MI attacks use statistical features derived from the model's outputs. These statistical features essentially represent an overall numerical abstraction of the model's predictions. In contrast, PartCrop utilizes a subtle part-aware capability that is implicitly developed during the training process of self-supervised learning (SSL), primarily focusing on sensitivity to and understanding of local information in an image.

\textbf{Challenges compared to attacking supervised models.} There may be three challenges in implementing an attack on a self-supervised model compared to a supervised one. Firstly, SSL allows the model to learn from substantial unlabeled data, avoiding severe overfitting on training data, which makes inferring member data harder. Second, SSL's proxy tasks~\cite{geiping2023cookbook} are richer and more sophisticated than merely fitting labels in supervised learning (SL). Therefore, a unified method is necessary. Finally, SSL's membership features require crafted designs and are not as straightforward as in SL, where features such as entropy/confidence can be used.

\textbf{Limitation and improvement.} For simplicity, PartCrop randomly crops patches from image to potentially yield part-containing crops with pre-define crop scale for all datasets. This may cause two potential issues: Random cropping could involve background noises, namely, patches without part of an object (noisy queries); Pre-define crop scale may not be generally suitable for all datasets whose image sizes are different. Consequently, these issues could lead to some queries of low quality. Considering that part is critical for attcking a self-supervised model, we use large number of crops in PartCrop to provide sufficient information to train attacker. 
Though this strategy in PartCrop is effectively verified by our experiments, it still wastes the computational resources to some extent due to noisy queries. To alleviate these issues, we could adopt an adaptive clipping strategy. For example, given an image, before clipping it, we can leverage a well-trained detector~\cite{carion2020end, liu2023grounding, wang2023detecting} to generate object bounding boxes as the approximate clipping scopes. Also, we can leverage a well-trained segmentor~\cite{kirillov2023segment, zhu2021crf, ma2024segment} to provide object masks that are more precise than bounding boxes. Regardless, these preprocessing methods effectively eliminate background noise, leading to more precise clipping areas compared to random clipping. Then we can crop object parts from these areas, by which we ensure that the cropped parts are relevant and contain meaningful content. This adaptive clipping strategy could significantly enhance the quality of the resulting parts. Additionally, we can further develop selection criteria for the clipping ratio for different datasets based on the results of these preprocessing methods. Specifically, we can compute the average size of all objects in the dataset. If the average size is small, we should appropriately lower the upper bound of the clipping ratio to prevent the cropped parts from being too large. Conversely, we can increase its lower bound. When combining these improvements together, we can refine the clipping process and ensure that the generated parts are optimized for various datasets, which enables us to consistently produce high-quality, meaningful parts that enhance the performance and training efficiency of PartCrop.

\section{Conclusion}
In this paper, we propose PartCrop, a unified method, to perform membership inference on a self-supervised model that could be trained by different training paradigms, without knowing how the self-supervised model is trained. This setting is more likely to meet the reality that an adversary usually have no knowledge about the training details of the self-supervised model as it is generally preserved in a black box. Specifically, by measuring the response between image and cropped parts in representation space, PartCrop produces disriminative membership feature for inference. Moreover, to defend against PartCrop, two common defense methods, \ie, early stop and differential privacy, are evaluated. We also propose a tailored defense method called shrinking crop scale range. Extensive experiments are conducted using three computer vision datasets. The results verify the effectiveness of the attack and defense methods.

\section{Acknowledgement}

We thank the anonymous reviewers for constructive comments. We also thank Ruiqing Ding, Chung-ju Huang, and Zhihao Yu for help and discussion. This work was partly supported by
 the National Science and Technology Major Project of China (2022ZD0119103).

\bibliographystyle{ACM-Reference-Format}
\bibliography{acmart}

\newpage

\appendix

\section{Appendix} ~\label{app:more results}

\begin{figure*}[h]	\centering{\includegraphics[width=1\linewidth]{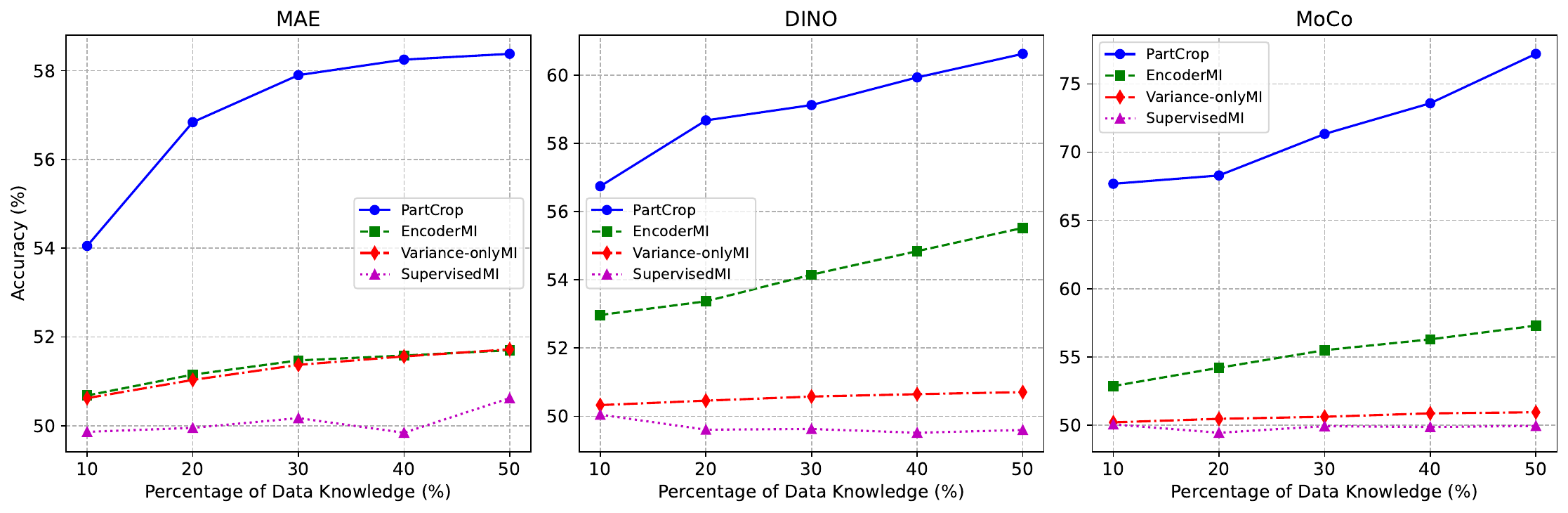}}
	\caption{The illustration of a sweep experiment on the adversary's knowledge assumption from 10\% to 50\%.}
	\label{fig:ratio}
 \vskip -0.05in
\end{figure*}

\textbf{Varying Adversary's Knowledge.} To demonstrate that PartCrop can work with limited knowledge about the samples, we perform an experiment using CIFAR100, varying the adversary's knowledge assumption from 10\% to 50\% on MAE, DINO, and MoCo, respectively, and compare the results with three baselines. As shown in Fig~\ref{fig:ratio}, the accuracy of PartCrop, EncoderMI, and Variance-onlyMI shows a steady improvement, while that of SupervisedMI varies irregularly. This is reasonable, as SupervisedMI is designed for supervised models and only performs random guesses in the self-supervised case. We also observe that PartCrop consistently outperforms EncoderMI and Variance-onlyMI under various settings of the adversary's knowledge. Moreover, though with only a 10\% knowledge assumption, PartCrop still achieves comparable accuracy (\eg, 56.74\% Acc on DINO), which outperforms that of EncderMI with a 50\% knowledge assumption (55.52\% Acc on DINO). This result highlights PartCrop's effectiveness and practicality in real-world scenarios.

\textbf{PartCrop's complementarity to other methods.} 
To validate this, we combine the membership features of PartCrop and EncoderMI to train an attacker to attack DINO using CIFAR100. This approach yields an accuracy of 62.02, further improving upon PartCrop's performance (60.62 Acc), which indicates that PartCrop is also complementary to other methods.

\textbf{Evaluation on diverse datasets.}
We evaluate PartCrop on more diverse and dynamic datasets, such as medical images, to better illustrate the effectiveness of the proposed method in various real-world scenarios. Specifically, we evaluate PartCrop on DINO, trained using the HAM10000 medical dataset~\cite{DVN/DBW86T_2018} with default settings. PartCrop achieves 56.32 Acc, outperforming EncoderMI (52.44 Acc), the strongest baseline, by a large margin. This result verifies the effectiveness and adaptability of PartCrop on diverse datasets.

\textbf{Runtime on high-resolution images.}
In real-world scenarios, images are often of high resolution. Hence, we adopt the widely-used ImageNet1K~\cite{deng2009imagenet, alexnet_krizhevsky2012imagenet} dataset with a 224x224 resolution on a large ViT-B model to evaluate PartCrop's inference runtime. The experiment is conducted on an NVIDIA 3090 (24GB) with an Intel Core i9-12900K. PartCrop's runtime is 67.8 ms per sample, faster than EncoderMI, the strongest baseline (107.8 ms per sample).

\end{document}